\documentclass[10pt,twocolumn,letterpaper]{article}

\usepackage[pagenumbers]{cvpr} % To force page numbers, e.g. for an arXiv version

\usepackage{algorithm}
\usepackage{algpseudocode}

\definecolor{cvprblue}{rgb}{0.21,0.49,0.74}
\usepackage[pagebackref,breaklinks,colorlinks,allcolors=cvprblue]{hyperref}

\def\confName{CVPR}
\def\confYear{2027}

\title{SSF-GS: Steady-State Filtered Densification for Efficient 3D Gaussian Splatting}

\author{Yongjae Lee\\
Arizona State University\\
Tempe, AZ, 85281\\
{\tt\small ylee298@asu.edu}
\and
Abhay Kumar Yadav\\
Johns Hopkins University\\
Baltimore, MD, 21218\\
{\tt\small ayadav13@jh.edu}
\and
Rama Chellappa\\
Johns Hopkins University\\
Baltimore, MD, 21218\\
{\tt\small rchella4@jhu.edu}
\and
Deliang Fan\thanks{Corresponding author}\\
Arizona State University\\
Tempe, AZ, 85281\\
{\tt\small dfan@asu.edu}
}

\begin{document}
\maketitle

\begin{abstract}
Adaptive density control in 3D Gaussian Splatting (3DGS) allocates scene
capacity by splitting or cloning Gaussians whose recent gradients exceed
a threshold. However, a gradient signal that temporarily exceeds the
densification threshold does not necessarily indicate a persistent need for
additional capacity, and acting on such signals can create unnecessary
Gaussians and increase training cost.
We introduce SSF-GS, a plug-and-play steady-state
filter (SSF) that withholds the backbone's density-control decisions while
signals remain unsettled.
SSF combines signal smoothing with normalized consecutive-window validation,
allowing Gaussian parameters to be further optimized so that high gradient
signals more reliably indicate a persistent rather than temporary need for
additional capacity.
This suppresses unnecessary intermediate Gaussian generation, reducing the
final Gaussian count and memory pressure.
Results across MipNeRF360, Tanks\&Temples, and Deep Blending
show that SSF reduces the final Gaussian count by 17.9\%--30.6\%, peak GPU
memory by 8.5\%--11.8\%, and training time by 6.2\%--14.3\% relative to
vanilla 3DGS, with comparable or higher PSNR.
SSF also complements existing efficient training methods, providing 
% up to 12.4\% 
additional training-time reduction with modest rendering quality change
on accelerated backbones, 
e.g., enabling FastGS+AdpSplit+SSF to train on MipNeRF360 in 250 seconds
($-82.1\%$ relative to vanilla 3DGS's 1,400 seconds).
Extensions to dynamic and sparse-view reconstruction demonstrate
its applicability to methods that employ signal-based densification across
diverse training settings.
\end{abstract}

\section{Introduction}
% Some quick background introduction in novel view synthesis and 3d gaussian splatting.
3D Gaussian Splatting (3DGS)~\cite{Kerbl20233DGaussian} has become a leading approach to
novel view synthesis and 3D scene reconstruction, supporting efficient and accurate
scene representations for applications such as simultaneous localization and mapping
(SLAM)~\cite{Keetha2024SplaTAM,Matsuki2024Gaussian,Yan2024GSSLAM,
Yugay2024GaussianSLAM,Sandström2025SplatSLAM}, robot manipulation~\cite{Lu2025ManiGaussian,Qureshi2025SplatSim,
Barcellona2025Dream,Shorinwa2025SplatMOVER,Yang2025Novel}, dynamic scene
reconstruction~\cite{Wu20244DGaussian,Yang2024Deformable,Luiten2024Dynamic,
Li2024Spacetime,Yang2024Realtime}, and physics-based
simulation~\cite{Xie2024PhysGaussian,Zhang2025PhysDreamer,Zhong2025Reconstruction,
Liu2024Physics3D}. Alongside anisotropic Gaussian primitives
and tile-based differentiable rasterization, 3DGS employs adaptive density control (ADC),
a key component that adjusts the number and spatial distribution of Gaussian primitives,
to achieve high-fidelity rendering at real-time frame rates.

% A short introduction adaptive density control and related works.
The intuition behind ADC is that Gaussians are likely to exhibit large view-space
position-gradient magnitudes in regions with insufficient representational capacity.
Vanilla ADC selects candidate Gaussians based on their recent gradient signals and increases
the capacity through multiple densification rounds, each involving splitting or cloning operations.
Following this intuition, recent work revises ADC 
to improve candidate Gaussian selection~\cite{Ren2025FastGS,Deng2026Improving,Ning2026Beyond}
or reduce the number of densification rounds needed~\cite{Lee2026AdpSplit}.
% such as multi-view consistent candidate selection~\cite{Ren2025FastGS},
% image edge aware candidate selection~\cite{Deng2026Improving},
% reconstruction-error-driven splitting~\cite{Lee2026AdpSplit}, and learned
% density-control policies~\cite{Ning2026Beyond}.

% Analysis of gaussian gradient with figs/fig2a.pdf and figs/fig2b.pdf, which is from
% mipnerf360/bicycle scene.
% Under fixed 1000th iter delay setting, trace the ratio of gaussians that become
% cross-threshold among set of above-threshold gaussians.
% → Most/Many gaussians (type what percentage does according to the figure) are become
% below-threshold until age of 1000.
% → This supports early exceedance doesn't necessarily represent a persistent
% densification request.
However, these ADC revisions leave an important question insufficiently addressed:
whether the currently observed gradient signal provides a reliable basis for
densification candidate selection. A large gradient can reflect ongoing adaptation of
an existing primitive rather than a persistent need for additional capacity.

\begin{figure*}[t]
    \centering
    \resizebox{.8\linewidth}{!}{%
    \begin{subfigure}[t]{0.49\textwidth}
        \centering
        \includegraphics[width=\linewidth]{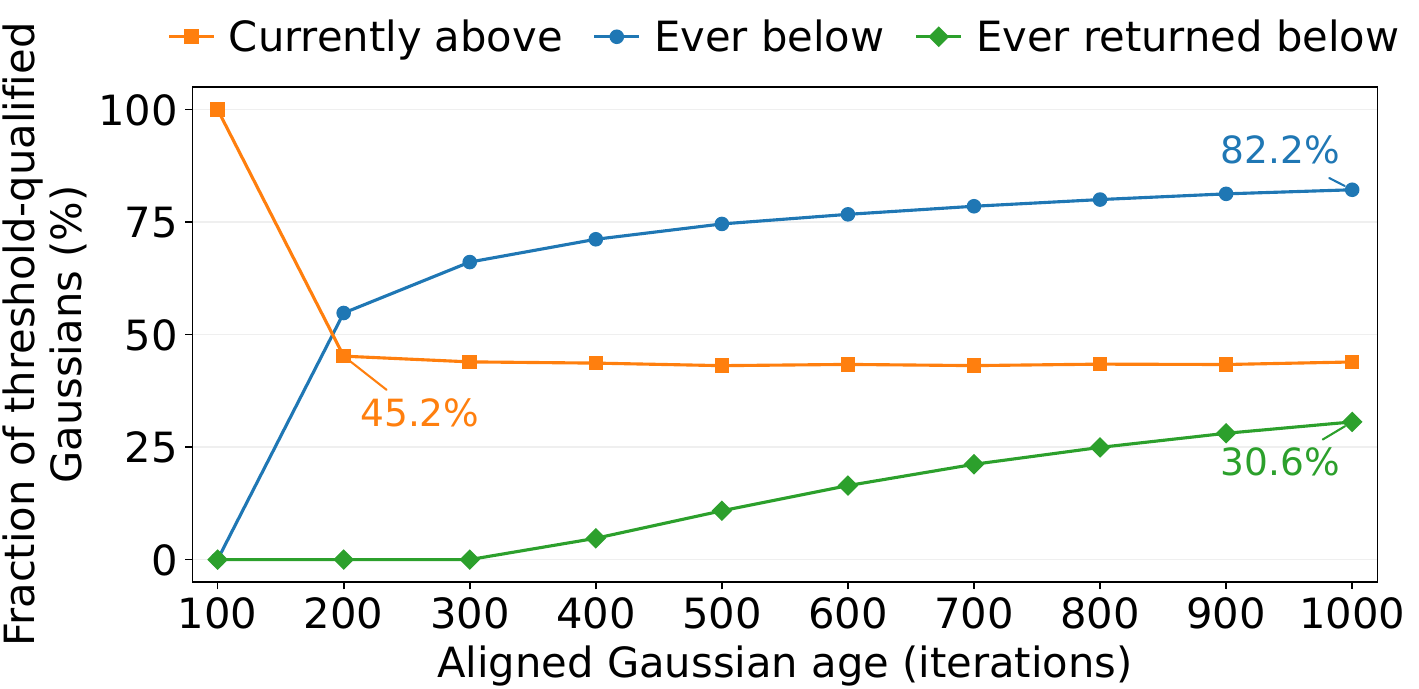}
        \caption{Gradient threshold crossings over aligned Gaussian age.}
        \label{fig:crossings}
    \end{subfigure}\hfill
    \begin{subfigure}[t]{0.49\textwidth}
        \centering
        \includegraphics[width=\linewidth]{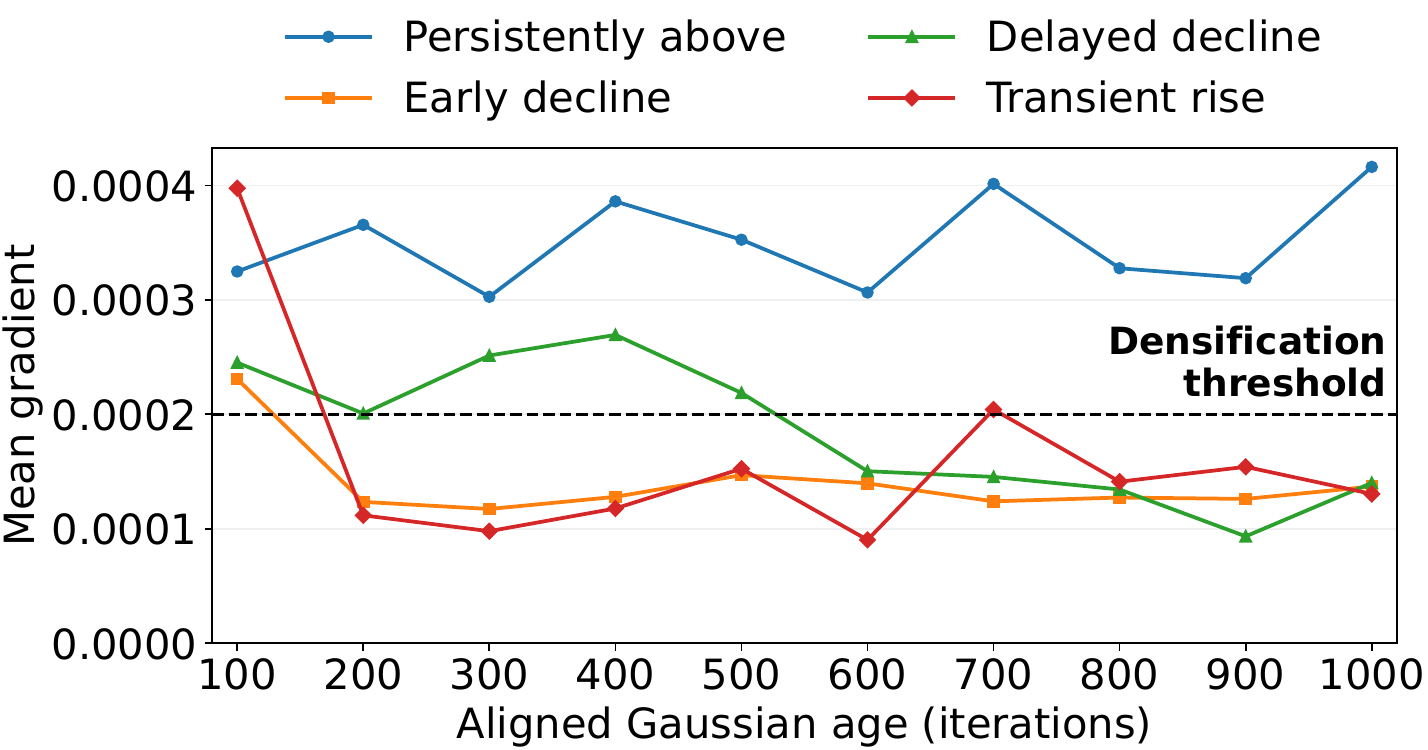}
        \caption{Example gradient histories.}
        \label{fig:histories}
    \end{subfigure}
    }
    \caption{\textbf{Threshold exceedance can be transient.}
    A Gaussian is \emph{threshold-qualified} when its mean position-gradient magnitude
    over a densification interval ($T_d=100$) meets or exceeds the densification threshold.
    (a) We track Gaussians that become threshold-qualified in the \texttt{Bicycle}
    scene from MipNeRF360~\cite{Barron2022Mipnerf360}, aligning each Gaussian's first
    qualification to age 100, where the ``Currently above'' fraction is 100\% by
    definition. We record whether their gradient signals remain above or fall below the threshold over 
    the following nine densification rounds while protecting them from ADC operations.
    (b) Selected recorded histories illustrate gradient signals that remain above the
    threshold, early and delayed declines, and a transient rise followed by
    consistently below-threshold gradients.
    The supplementary material provides details of the setup of this experiment.}
    \label{fig:grad_analysis}
\end{figure*}

% We refer to a Gaussian as eligible for densification when its mean gradient magnitude is at or
% above the densification threshold. Eligibility in a single interval alone does not
% establish that the refinement request will persist.
% To examine this issue, we track initially above-threshold Gaussians in the Bicycle scene from
% MipNeRF360~\cite{Barron2022Mipnerf360} under delayed densification.
A trace on threshold-qualified Gaussians reveals that initial threshold qualification
does not necessarily indicate a persistent need for densification.
As shown in \autoref{fig:crossings}, only 45.2\% of the initially threshold-qualified Gaussians
remain above the threshold in the next densification round
(``Currently above''), and 82.2\% of the same cohort fall below it
at least once during the following nine rounds (``Ever below'').
Moreover, 30.6\% have fallen
below the threshold again at least once after a rebound above it (``Ever returned below'').
% → the 4 example gaussian gradient histories show that the grad signal include
% transient threshold crossings before its signal stable.
% → Intuitively determining the densification operation based on stabilized gradient
% signal, not on transient gradient signal.
% → Thus this observation motivates us to use steady state filter for no transient
% gradient gaussian selection.
The mean gradient histories in \autoref{fig:histories} illustrate how gradient signals
vary over time. Among the four signals that initially satisfy the same threshold criterion, 
“Persistently above” consistently remains above the threshold, 
while the others eventually fall below it as parameter optimization proceeds.
In particular,
``Transient rise'' returns below the threshold by aligned age 800 shortly after
exceeding it again, highlighting that the recent gradient signal alone cannot
reliably indicate whether its magnitude will remain above the threshold or
fall below it again. Densification decisions based solely on the recent signal
may therefore introduce unnecessary primitives.

% We propose xxx (a proposed method name).
To avoid unnecessary densification caused by signals that only temporarily
exceed the threshold, we propose \textbf{SSF-GS}, a steady-state filter (SSF) that
determines when a Gaussian is eligible for ADC.
The filter smooths each Gaussian's gradient signal and applies normalized
consecutive-window (NCW) validation to its changes at regular intervals.
% Each change is measured
% relative to the largest smoothed value recorded for that Gaussian. 
When the changes remain sufficiently small for several consecutive checks,
the Gaussian becomes eligible for selection under the host method's existing
gradient threshold and topology rules.
SSF-GS thus assesses whether the
signal has settled before testing whether its magnitude warrants densification.
By reducing unnecessary densification driven by transiently high gradients,
SSF-GS achieves a smaller Gaussian representation, lowers peak GPU memory
usage, and accelerates training.

% Our contribution is:
% 1. simple and general approach: the proposed nomalized consecutive window (NCW) filter
% (a type of steady-state filter)
% approach simple enough to apply gradient signal based densification methods.
% 2. This NCW filter is better than other filters in the steady-state filter class. Ours
% shows the best Gaussian--rendering quality tradeoff.
% 3. Generalizablilty: This NCW filter is compatible with SOTA densification methods,
% further improving efficiency.
Our contributions are threefold:
\begin{itemize}
    \item We introduce SSF-GS, a lightweight, plug-and-play steady-state filter using
    NCW validation to reduce unnecessary densification and accelerate training.
    \item We show that SSF achieves a better balance between Gaussian count
    and rendering quality than existing reliability-based Gaussian selection methods.
    \item Integration of SSF into state-of-the-art 3DGS training accelerators
    demonstrates its generalizability through further improvements in training efficiency.
\end{itemize}

\section{Related Work}
\label{sec:related}
% subsection 1: online steady state filter
% (reviews papres realted traditional signal analysis)
% - Songling Cao and R.Russell Rhinehart, An efficient method for on-line identification
% of steady state, Journal of Process Control, 1995
% - Jeffrey D. Kelly and John D. Hedengren, A steady-state detection (SSD) algorithm to
% detect non-stationary drifts in processes, Journal of Process Control, 2013
% - Kim et al., Design of a steady-state detector for fault detection and diagnosis of a
% residential air conditioner, International Journal of Refrigeration, 2008
\subsection{Online Steady-State Detection}
In systems theory, steady-state detection aims to identify when a signal reaches a
condition in which its statistical properties remain approximately constant, 
so that subsequent analysis and decisions are based on stable measurements.
Unlike offline detection~\cite{Rhinhart2022Applied}, 
which can examine a complete recorded sequence, online
detection must assess stability as measurements arrive without access to future
observations, balancing timely detection against the risk of mistaking a temporary
plateau for a steady state~\cite{Cao1995SteadyState}.
Cao and Rhinehart~\cite{Cao1995SteadyState} compared smoothed squared
deviations from a filtered mean with those between consecutive measurements
using an F-like statistic to distinguish mean drift from short-term fluctuations.
Kelly and Hedengren~\cite{Kelly2013SteadyStateDetection} used a residual
Student-$t$ test to assess departures from a constant mean, while
Kim et al.~\cite{Kim2008Design} measured variability over moving windows
to limit the influence of past transients.
These approaches provide a principled basis for online steady-state detection,
which motivates our investigation of signal stability before densification.

% subsection 2: reliable gaussian candidate
% (reviews papers that use some mechanisms to select reliable gaussisns to be densified)
% - Temporally Aware Densification for Dynamic 3D Gaussian Splatting, ECCV 2026
% - Improving 3D Gaussian splatting with reliability-guided densification and
% hierarchical semantic supervision, JVCIR 2026
% - Reliability-Guided Densification for 3D Gaussian Splatting, preprint 2026 May
% - CAdam: Context-Adaptive Moment Estimation for 3D Gaussian Densification in
% Generative Distillation, SIGGRAPH 2026
\subsection{Reliable Evidence for Gaussian Densification}
Because the recent gradient signal alone does not provide reliable
evidence for densification, recent methods incorporate additional information to
improve the reliability of candidate selection.
This information includes temporal visibility support~\cite{Sandu2026Temporally}
and observation reliability~\cite{Pang2026Improving}.
These criteria improve the evidence for candidate selection but do not directly
establish whether the signal has stabilized, potentially leaving
densification decisions sensitive to transient changes.
CAdam~\cite{Chung2026CAdam} uses a signal-to-noise (SNR) ratio derived from the first and
second moments of world-space positional gradients to assess their directional consistency.
However, directional consistency does not necessarily imply a stable signal
magnitude, as signals can remain aligned while their magnitudes change.

% subsection 3: other densification methods
% (reviews papers that perpendicular to our approach in terms of training efficiency)
% - FastGS: Training 3D Gaussian Splatting in 100 Seconds, CVPR 2026
% - Improving Densification in 3D Gaussian Splatting for High-Fidelity Rendering, CVPR
% 2026
% - Beyond Heuristics: Learnable Density Control for 3D Gaussian Splatting, ICML 2026
% - AdpSplit: Error-Driven Adaptive Splitting for Faster Geometry Discovery in 3D
% Gaussian Splatting, Arxiv
% - Speedy-Splat: Fast 3D Gaussian Splatting with Sparse Pixels and Sparse Primitives,
% CVPR 2025
% - DashGaussian: Optimizing 3D Gaussian Splatting in 200 Seconds, CVPR 2025
\subsection{Efficient 3D Gaussian Splatting Training}
Recent methods improve training efficiency by allocating Gaussian capacity
through targeted densification.
FastGS~\cite{Ren2025FastGS} guides densification and pruning with multi-view
consistency, while ImprovedGS~\cite{Deng2026Improving} combines edge-aware
selection with long-axis splitting to preserve details and limit distortion.
LeGS~\cite{Ning2026Beyond} learns density-control policies through reinforcement
learning with reconstruction-quality rewards. AdpSplit~\cite{Lee2026AdpSplit}
uses image-error statistics to determine split-child counts and initialization,
reducing the densification rounds needed to discover underlying geometry.
Other methods accelerate training by pruning redundant
primitives~\cite{Hanson2025SpeedySplat,Liu2025EfficientGS}, scheduling Gaussian
growth~\cite{Chen2025DashGaussian,Mallick2024Taming3DGS,Deng2026Improving}, or
replacing Adam with a tailored Levenberg--Marquardt solver~\cite{Höllein20253DGSLM}.
Our filter approach complements these strategies by determining when Gaussians become
eligible for densification while preserving the backbone's selection criteria
and topology operators.

\section{SSF-GS}
\label{sec:method}
\begin{figure}[t]
    \centering
    \includegraphics[width=.7\linewidth]{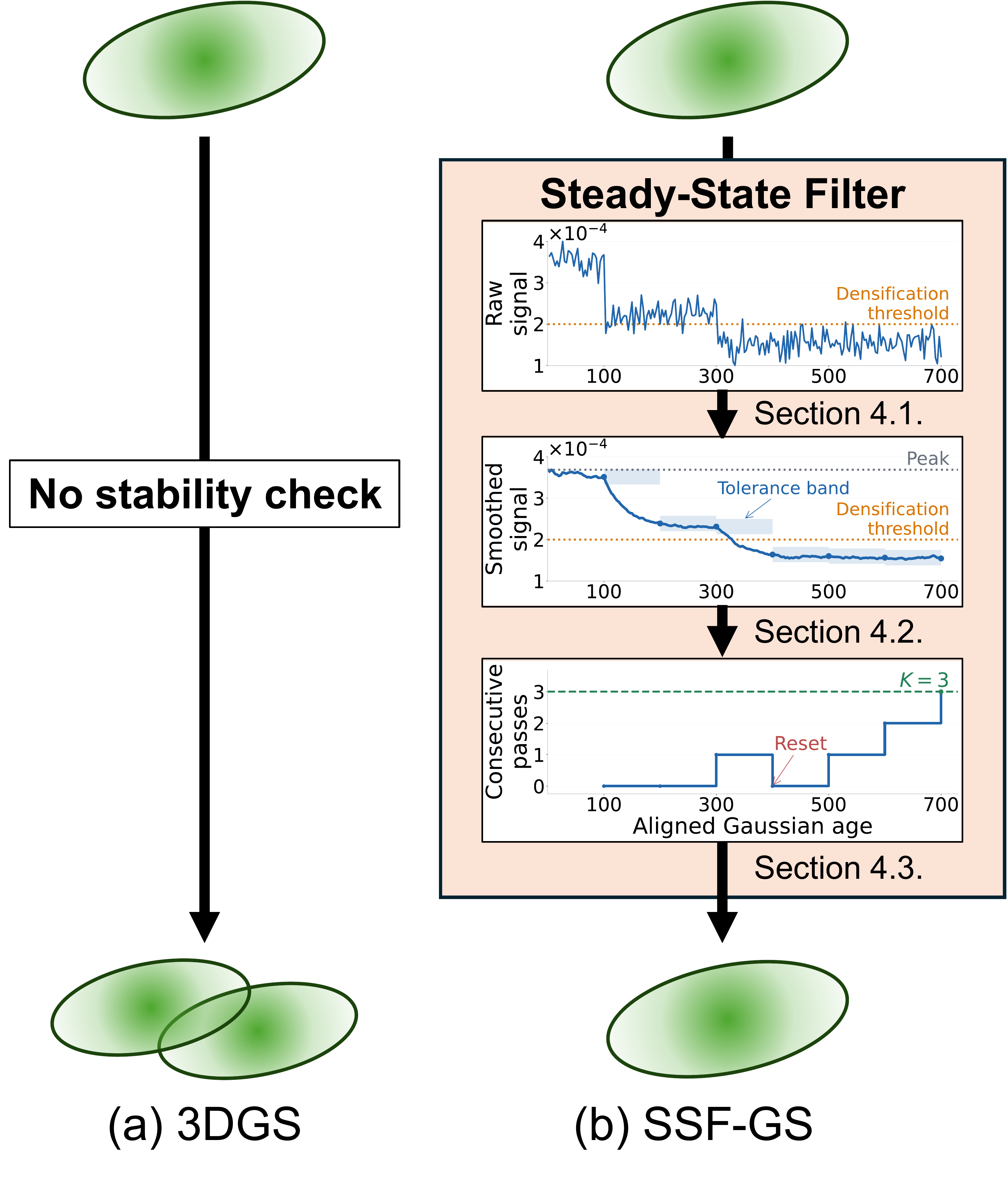}
    \caption{\textbf{Overview of SSF-GS.}
    (a) Without a stability check, signals that temporarily exceed the
    threshold can trigger unnecessary densification.
    (b) Our steady-state filter first smooths the observed signal,
    evaluates its normalized change against a settling tolerance
    band, and releases the Gaussian when it reaches consecutive pass condition.
    In this example, the signal settles below the densification threshold,
    so the Gaussian remains undensified after release, avoiding additional, redundant primitives.}
    \label{fig:ncwv-overview}
\end{figure}

We introduce SSF-GS, which incorporates a steady-state filter (SSF) to reduce the
influence of transient signals on densification decisions. Through normalized
consecutive-window (NCW) validation, the filter assesses signal stability before releasing
a Gaussian for the backbone's topology operations (\autoref{fig:ncwv-overview}).

\subsection{Preliminary: 3D Gaussian Splatting}
\label{sec:preliminary}
3D Gaussian Splatting (3DGS)~\cite{Kerbl20233DGaussian} represents a scene as a set of
Gaussian primitives initialized from a sparse point cloud.
Each Gaussian $i$ is parameterized by its position $\boldsymbol\mu_i$, scale
$\mathbf{s}_i$, rotation quaternion $\mathbf{q}_i$, opacity $o_i$, and spherical-harmonic
coefficients defining its view-dependent color $\mathbf{c}_i$.

For a given camera, 
the differentiable rasterizer composites the ordered set of Gaussians
$\mathcal{N}(\mathbf{p})$ contributing to pixel $\mathbf{p}$ in front-to-back depth
order:
\begin{equation}
    \mathbf{C}(\mathbf{p})=
    \sum_{i\in\mathcal{N}(\mathbf{p})}\mathbf{c}_i\alpha_i(\mathbf{p})
    \prod_{\substack{j\in\mathcal{N}(\mathbf{p})\\j<i}}\bigl(1-\alpha_j(\mathbf{p})\bigr),
    \label{eq:rendering}
\end{equation}
where $j<i$ denotes contributors preceding $i$ in this depth order,
and $\alpha_i(\mathbf{p})$ is the opacity $o_i$ scaled by the projected Gaussian's
value at pixel $\mathbf{p}$.
The Gaussian parameters are optimized using the photometric objective:
\begin{equation}
    \mathcal{L}=(1-\lambda)\mathcal{L}_1+\lambda\mathcal{L}_{\mathrm{D\text{-}SSIM}}.
\end{equation}
Here, $\lambda$ balances the pixel-wise $\mathcal{L}_1$ loss and structural
dissimilarity between the rendered and ground-truth images.

Let $t$ denote the training iteration and $T_d$ the densification interval
in iterations. Adaptive density control (ADC) runs every $T_d$ iterations
during the densification phase and averages view-space positional gradient
magnitudes over visible observations between rounds:
\begin{equation}
    \begin{aligned}
        g_i(t) &=
        \left\|\frac{\partial\mathcal{L}_t}
        {\partial\boldsymbol\mu_i^{\mathrm{2D}}}\right\|_2,\\
        \bar g_i(t) &=
        \frac{1}{|\mathcal{W}_i|}
        \sum_{u\in\mathcal{W}_i}g_i(u).
    \end{aligned}
    \label{eq:adc}
\end{equation}
Here, $\mathcal{L}_t$ is the photometric loss at iteration $t$, and
$\boldsymbol\mu_i^{\mathrm{2D}}$ is the two-dimensional location of Gaussian $i$'s
projected center in the rendered image. For the current densification interval
of length $T_d$, $\mathcal{W}_i$ denotes the iterations in which Gaussian $i$
is visible. ADC sets $\bar g_i(t)=0$ when this set is empty.

At each round, the threshold-qualified and pruning sets are
\begin{equation}
    \begin{aligned}
        \mathcal{D}(t) &=
        \{i:|\mathcal{W}_i|>0,\ \bar g_i(t)\geq\tau_g\},\\
        \mathcal{P}(t) &= \{i:o_i<\tau_o\},
    \end{aligned}
    \label{eq:vanilla-selection}
\end{equation}
where $\tau_g>0$ is the densification threshold and $\tau_o$ is the minimum opacity.
Gaussians in $\mathcal{D}(t)$ whose largest scale is at most the predefined
threshold $\tau_s$ are cloned by adding a copy while retaining the parent,
whereas those with larger scales are split by replacing the parent with
smaller children.
These operations adapt scene capacity alongside continuous parameter optimization.

% As illustrated in , the filter comprises three stages.
% First, we smooth valid observations to reduce short-term fluctuations
% (\autoref{sec:smoothing}). Second, we measure changes between consecutive check
% boundaries relative to the signal's historical filtered peak
% (\autoref{sec:normalized-change}). Third, we require consecutive changes to remain
% within a tolerance before confirming stability (\autoref{sec:ncw}).
% Release allows the backbone to evaluate its own densification criterion and does
% not automatically trigger densification.
% \autoref{alg:ncwv} gives the complete filtering procedure in the supplement.

\subsection{Signal Smoothing}
\label{sec:smoothing}
We model the raw signal used to assess insufficient capacity for Gaussian $i$
as an underlying response $x_{i,n}$ plus view-dependent noise $\eta_{i,n}$:
\begin{equation}
    g_{i,n}=x_{i,n}+\eta_{i,n},\qquad n>0,
    \label{eq:signal-model}
\end{equation}
where $n$ counts finite, visible observations since Gaussian $i$ was created,
and $g_{i,n}$ denotes $g_i(t)$ at its $n$-th valid observation. 

The underlying response varies as the Gaussian's position, scale, opacity,
and appearance are optimized. Different sampled views also introduce variation
in observations of the same Gaussian. To reduce sensitivity to this
view-dependent variation, we smooth the signal before assessing changes in
its accumulated response. Specifically, we use an exponential moving average
(EMA) accumulator $m_{i,n}$ and correct
its zero-initialization bias to obtain the smoothed signal
$\hat{m}_{i,n}$~\cite{Kingma2017Adam}:
\begin{equation}
    \begin{aligned}
        m_{i,n} &= \beta m_{i,n-1}+(1-\beta)g_{i,n},\\
        \hat{m}_{i,n} &= \frac{m_{i,n}}{1-\beta^n}.
    \end{aligned}
    \label{eq:ema}
\end{equation}
The decay factor $\beta$ controls the weight retained
from past observations, with larger values providing stronger smoothing.

The EMA advances only when the Gaussian has a finite, visible observation.
An unobserved Gaussian therefore retains its previous state rather than
receiving a zero-valued sample, which would incorrectly suggest a declining
signal. Bias correction prevents newly created Gaussians from appearing to
have small responses solely because their accumulators start at zero.

\subsection{Normalized Signal Change}
\label{sec:normalized-change}
Signal magnitudes vary across Gaussians, making absolute changes unsuitable
for comparing their stability. We therefore normalize changes by each
Gaussian's historical peak, which provides a persistent reference scale as
its response decreases during optimization. We compute this peak from the
smoothed signal:
\begin{equation}
    p_{i,n}=\max_{j\leq n}\hat{m}_{i,j}.
    \label{eq:peak}
\end{equation}
Using $\hat{m}_{i,j}$ rather than $g_{i,j}$ reduces the influence of
view-dependent spikes that could inflate the normalization factor and make
subsequent signal changes appear artificially small.

% The peak is updated on every valid observation. 
% The smoothed signal $\hat{m}_{i,n}$
% and its peak $p_{i,n}$ are maintained across ADC's periodic accumulator resets.
% Updating this peak throughout the Gaussian's lifetime also accommodates
% increases beyond its earlier response range.

We update the peak on every valid observation and retain both the smoother
and peak across ADC's accumulator resets. Stability is therefore evaluated
using lifetime statistics, while threshold qualification uses the backbone's
recent interval-wise observations.

Let $a_i(t)$ denote the number of iterations since Gaussian $i$ was created.
We check stability every $T_d$ iterations in alignment with ADC's schedule,
before evaluating threshold qualification. Comparisons begin once
$a_i(t)\geq a_{\mathrm{warm}}$ and require both a finite smoothed signal and
at least one new valid observation since the previous check. At the first
check satisfying these conditions, we store the smoothed signal as a reference
without registering a pass. If the conditions also hold at the next check,
we compare the current smoothed signal $\hat{m}_{i,n}$ with the reference
$\hat{m}_i^{\mathrm{prev}}$ stored at the previous check.
\begin{equation}
    d_i=\frac{|\hat{m}_{i,n}-\hat{m}_i^{\mathrm{prev}}|}{p_{i,n}}.
    \label{eq:drift}
\end{equation}
A comparison passes when $d_i(t)\leq\epsilon_{\mathrm{tol}}$, where $\epsilon_{\mathrm{tol}}$ is the
settling tolerance, with $\epsilon_{\mathrm{tol}}\geq0$.

\subsection{Consecutive-Window Validation}
\label{sec:ncw}
Smoothing reduces observation-level fluctuations, but a single small change
can still reflect a temporary plateau. We therefore validate settling over
consecutive comparisons. Let $s_i(t)$ count consecutive checks within the
settling tolerance. A qualified check satisfying $d_i\leq\epsilon_{\mathrm{tol}}$
increments the counter, while a failed check resets it. Each comparison
updates $\hat{m}_i^{\mathrm{prev}}$ to the current smoothed signal, regardless
of its outcome. Once $s_i(t)\geq K$, we set and retain the Boolean flag
$\mathrm{settled}_i=\mathrm{true}$.

Because the reference advances at each comparison, validation assesses
successive changes rather than agreement with a fixed initial value.
It therefore identifies local settling of the smoothed response rather than
convergence of the Gaussian parameters, since small successive changes can
still accumulate over time.

A signal that continues increasing above the densification threshold or
oscillates substantially may not settle within the training period.
To avoid indefinitely withholding topology updates, we also release a Gaussian
when its age reaches $a_{\mathrm{timeout}}$. At each ADC round, the backbone
applies its topology operations to the filtered candidate sets
\begin{equation}
    \begin{aligned}
        \mathcal{D}'(t)&=\{i\in\mathcal{D}(t):e_i(t)=1\},\\
        \mathcal{P}'(t)&=\{i\in\mathcal{P}(t):e_i(t)=1\},
    \end{aligned}
    \label{eq:selection}
\end{equation}
where the topology eligibility mask is
\begin{equation}
    e_i(t)=\mathbb{1}\!\left[
        \mathrm{settled}_i\ \lor\ a_i(t)\geq a_{\mathrm{timeout}}
    \right],
    \label{eq:eligibility}
\end{equation}
and $\mathbb{1}$ denotes the indicator function.

Release permits the backbone to evaluate its original selection rule and
does not itself trigger densification. In particular, a released Gaussian
is cloned or split only if its ADC mean $\bar g_i(t)$ satisfies the
densification threshold, independently of the smoothed signal used to assess
stability. Applying the same eligibility mask to pruning gives newly created
Gaussians time to optimize before low opacity leads to removal.

Filter state follows the backbone's topology updates. Vanilla ADC removes
split parents but retains clone parents and their optimizer state.
Accordingly, clone parents retain their SSF statistics, including age,
settling state, and lifetime signal statistics, while new children start
with fresh filter state and removed Gaussians have their state discarded.

\begin{figure*}[t]
    \centering
    \includegraphics[width=\linewidth]{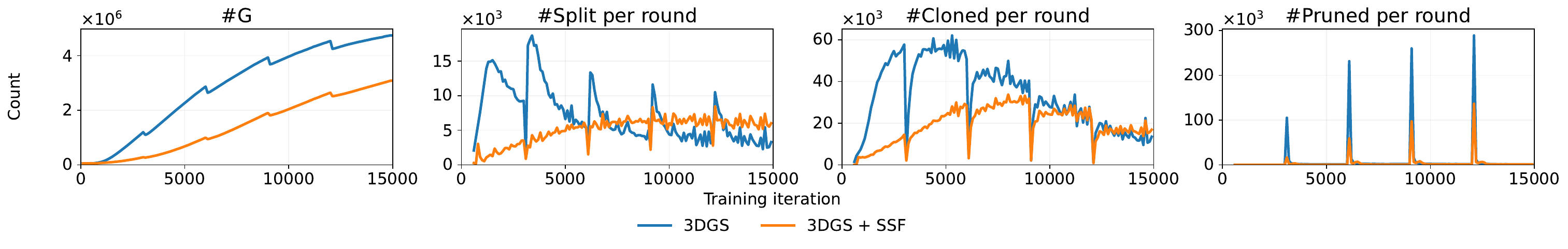}
    \caption{\textbf{Gaussian population and topology operations during the densification
    phase on \texttt{Bicycle} scene.} SSF blocks densification decision 
    on insufficiently optimized gaussians, reaching the smaller final gaussian count 
    at the densification phase end (15k), as the curves in \#Split and \#Cloned 
    show milder surge early stage of densification phase around 0-10k, compared to 3DGS.}
    \label{fig:fig4}
\end{figure*}

\section{Experiments}
\label{sec:experiments}

% subsection 1: setup
\subsection{Setup}

% paragraph 1: Datasets
% MipNeRF360, Tanks&Temples, and Deep-Blending.
\paragraph{Datasets.}
We evaluate on nine scenes from MipNeRF360~\cite{Barron2022Mipnerf360},
two from Tanks\&Temples~\cite{Knapitsch2017Tanks}, and two from
Deep Blending~\cite{Hedman2018Deep}. For these datasets, we follow the 3DGS
evaluation protocol, holding out every eighth image for evaluation and using
the remaining images for training. Images wider than 1,600 pixels are
downsampled to that width while preserving their aspect ratio.
For dynamic reconstruction, we evaluate on eight scenes from
HyperNeRF~\cite{Park2021HyperNeRF} and seven scenes from
NeRF-DS~\cite{Yan2023NeRFDS}. For sparse-view reconstruction, we use eight
scenes from LLFF~\cite{Mildenhall2019Local}, with nine training views per scene.

% paragraph 2: Metrics
\paragraph{Metrics.}
We assess reconstruction quality using PSNR (dB), SSIM, and LPIPS.
To measure efficiency, we report the final Gaussian count, cumulative
densification count (the sum of split and cloned parent counts), and
cumulative pruning count, all in millions. 
We also report total training time (seconds), rendering throughput
(frames per second), and peak GPU memory (GB).

% paragraph 3: Baselines
\paragraph{Baselines.}
We compare our steady-state filter with vanilla
3DGS~\cite{Kerbl20233DGaussian}, SNR gating from
CAdam~\cite{Chung2026CAdam}, reliable-observation gradient statistics (ROG)
from RDS-GS~\cite{Pang2026Improving}, and visibility-weighted gradients (VWG)
from TAD-GS~\cite{Sandu2026Temporally}.
% Reliability-Guided Densification (RGD) from RGD-GS~\cite{Zachari2026Reliability}
For a controlled comparison, we implement their signal-reliability-based
candidate selection mechanisms rather than
reproduce the complete systems, which include additional techniques.
\footnote{Source code for CAdam and RDS-GS was not publicly available
at the time of submission.}
% Implementation details of these baselines are provided in
% \autoref{sec:baseline-adaptations}.

We additionally evaluate SSF with absolute-gradient accumulation
(AbsGS)~\cite{Ye2024AbsGS}, which can identify densification candidates
missed by the signed-gradient threshold by reducing gradient cancellation.
% Both RGD and
% through image-error-based reliability scores and reduced gradient cancellation,
% respectively.
Comparing AbsGS with AbsGS+SSF tests whether SSF remains effective with
an alternative monitoring signal.

To assess whether SSF provides additional training acceleration, we integrate
it with ImprovedGS~\cite{Deng2026Improving}, LeGS~\cite{Ning2026Beyond},
FastGS~\cite{Ren2025FastGS}, and FastGS+AdpSplit~\cite{Lee2026AdpSplit}.
We further evaluate its applicability to dynamic reconstruction and sparse-view
training using Deform3DGS~\cite{Yang2024Deformable} and
DropGaussian~\cite{Park2025DropGaussian}, respectively.

% paragraph 4: Implementation detail
\paragraph{Implementation details.}
Comparisons on the shared backbone use the default 3DGS hyperparameters:
30k training iterations, densification every 100 iterations until iteration
15k, opacity resets every 3k iterations, and $\tau_g=0.0002$.
All experiments are conducted on an NVIDIA A100 GPU.

For SSF-GS, we set the EMA decay factor to $\beta=0.999$, the warmup age to
$a_{\mathrm{warm}}=100$, the timeout age to $a_{\mathrm{timeout}}=800$, the
settling tolerance to $\epsilon_{\mathrm{tol}}=0.05$, 
and required consecutive validation check pass count $K=3$. 
% Settling requires $K=3$
% consecutive comparisons within tolerance.
% Newly created children receive a new birth iteration, zeroed smoothing
% statistics, and a consecutive-pass counter $s_i$ initialized to zero.

SSF maintains constant-size state per Gaussian, requiring $O(N)$ storage
and computation per update for $N$ live Gaussians, without storing trajectories
or adding inference-time overhead.

% Source: 1. A100_3dgs_cadam_rds_tad_vs_ssf.xlsx; second sheet (dataset summaries).
\begin{table}[t]
    \centering
    \caption{Comparison of signal-reliability-oriented methods' results,
    averaged by dataset.
    Time is in seconds and PM. is peak GPU memory in GB.
    \#G. is the final Gaussian count. \#D (Densified) is the cumulative
    sum of split and cloned parent counts, and \#P (Pruned) is the cumulative
    pruned Gaussian count. All counts are in millions. Best values are bolded, except for \#D and \#P.}
    \label{tab:main}

    \setlength{\tabcolsep}{3pt}
    \resizebox{\linewidth}{!}{%
    \begin{tabular}{lrrrrrrrrr}
        \toprule
        Method & PSNR$\uparrow$ & SSIM$\uparrow$ & LPIPS$\downarrow$ &
        Time$\downarrow$ & FPS$\uparrow$ & PM.$\downarrow$ &
        \#G.$\downarrow$ & \#D & \#P \\
        \midrule
        \multicolumn{10}{l}{\textbf{\texttt{MipNeRF360}}~\cite{Barron2022Mipnerf360}} \\
        % Run: 3dgs_0
        3DGS~\cite{Kerbl20233DGaussian} & 27.526 & 0.8127 & 0.2208 & 1400 & 109.2 & 9.92 &
        2.64 & 3.13 & 0.61 \\
        % Run: 3dgs_cadam_full_0_th0.30
        3DGS + SNR~\cite{Chung2026CAdam} & 24.756 & 0.7689 & 0.2700 & 1489 & 80.9 & 9.62 &
        2.32 & 2.49 & 0.29 \\
        % Run: 3dgs_rds_0
        3DGS + ROG~\cite{Pang2026Improving} & 27.570 & \textbf{0.8211} & \textbf{0.2010} & 1733 & 83.7 & 11.65 &
        3.71 & 4.49 & 0.89 \\
        % Run: 3dgs_tad_0
        3DGS + VWG~\cite{Sandu2026Temporally} & 27.536 & 0.8128 & 0.2209 & 1407 & 108.2 & 9.92 &
        2.64 & 3.13 & 0.61 \\
        % Run: 3dgs_ncwv_ema_tol0.05_w100_tout800_0
        3DGS + SSF & \textbf{27.572} & 0.8108 & 0.2285 & \textbf{1292} & \textbf{118.8} & \textbf{8.75} &
        \textbf{1.83} & 2.02 & 0.30 \\
        \midrule
        \multicolumn{10}{l}{\textbf{\texttt{Tanks\&Temples}}~\cite{Knapitsch2017Tanks}} \\
        % Run: 3dgs_0
        3DGS~\cite{Kerbl20233DGaussian} & 23.758 & 0.8496 & 0.1708 & 777 & 138.8 & 4.73 &
        1.56 & 2.09 & 0.69 \\
        % Run: 3dgs_cadam_full_0_th0.30
        3DGS + SNR~\cite{Chung2026CAdam} & 21.667 & 0.8107 & 0.2018 & 791 & 113.1 & 4.66 &
        1.49 & 1.63 & 0.30 \\
        % Run: 3dgs_rds_0
        3DGS + ROG~\cite{Pang2026Improving} & 23.239 & \textbf{0.8529} & \textbf{0.1460} & 1617 & 60.3 & 8.87 &
        4.16 & 6.85 & 2.84 \\
        % Run: 3dgs_tad_0
        3DGS + VWG~\cite{Sandu2026Temporally} & 23.797 & 0.8495 & 0.1708 & 780 & 138.0 & 4.74 &
        1.57 & 2.10 & 0.69 \\
        % Run: 3dgs_ncwv_ema_tol0.05_w100_tout800_0
        3DGS + SSF & \textbf{24.107} & 0.8512 & 0.1722 & \textbf{729} & \textbf{148.4} & \textbf{4.33} &
        \textbf{1.28} & 1.59 & 0.47 \\
        \midrule
        \multicolumn{10}{l}{\textbf{\texttt{Deep Blending}}~\cite{Hedman2018Deep}} \\
        % Run: 3dgs_0
        3DGS~\cite{Kerbl20233DGaussian} & 29.701 & 0.9026 & 0.2408 & 1350 & 115.1 & 8.11 &
        2.46 & 3.31 & 0.91 \\
        % Run: 3dgs_cadam_full_0_th0.30
        3DGS + SNR~\cite{Chung2026CAdam} & 27.726 & 0.8919 & 0.2523 & 1316 & 92.9 & 9.95 &
        3.74 & 4.15 & 0.47 \\
        % Run: 3dgs_rds_0
        3DGS + ROG~\cite{Pang2026Improving} & 29.581 & 0.9012 & \textbf{0.2374} & 1764 & 87.6 & 10.32 &
        3.79 & 5.55 & 1.82 \\
        % Run: 3dgs_tad_0
        3DGS + VWG~\cite{Sandu2026Temporally} & 29.755 & 0.9028 & 0.2407 & 1366 & 114.5 & 8.11 &
        2.46 & 3.31 & 0.91 \\
        % Run: 3dgs_ncwv_ema_tol0.05_w100_tout800_0
        3DGS + SSF & \textbf{29.796} & \textbf{0.9042} & 0.2448 & \textbf{1157} & \textbf{132.1} & \textbf{7.40} &
        \textbf{1.99} & 2.48 & 0.55 \\
        \bottomrule
    \end{tabular}%
    }
\end{table}

% subsection 2: Results
\subsection{Results}

% subsubsection 1: Reliable gaussian selection
% Show the comparison between 3dgs, 3dgs+cadam, 3dgs+rds, 3dgs+tad, and 3dgs+ours.
% (use experiments.xlsx)
\paragraph{Reliable Gaussian Selection.}
\autoref{tab:main} compares SSF with vanilla 3DGS and
signal-reliability-oriented methods~\cite{Chung2026CAdam,Pang2026Improving,Sandu2026Temporally}.
% adaptations of 3DGS using dataset-averaged results.
On MipNeRF360, SSF reduces the Gaussian count by 30.6\% and training time
by 7.7\% relative to vanilla 3DGS, while increasing PSNR from 27.526 to
27.572\,dB. This slight PSNR gain accompanies modest perceptual-quality
losses, with SSIM decreasing from 0.8127 to 0.8108 and LPIPS increasing
from 0.2208 to 0.2285.

The population reduction stems from fewer topology updates.
As shown in \autoref{fig:fig4}, SSF suppresses early splitting and cloning
in \texttt{Bicycle}, reducing their cumulative counts from 1.03\,M and
4.80\,M to 0.73\,M and 2.79\,M, respectively. Pruning also falls from
1.14\,M to 0.49\,M, indicating that fewer Gaussians are created and
subsequently removed. However, the smaller population does not yield
proportional training-time savings because rendering cost also depends on
visibility and projected coverage. In \texttt{Bicycle}, the final population
decreases by 35.0\% and training time by 16.7\%, while visible Gaussian
counts after densification decrease by only 18.0\%. Larger projected
footprints could further explain this efficiency gap and the loss of fine
detail despite comparable pixel-wise fidelity.

\begin{figure*}[t]
    \centering
    \includegraphics[width=\linewidth]{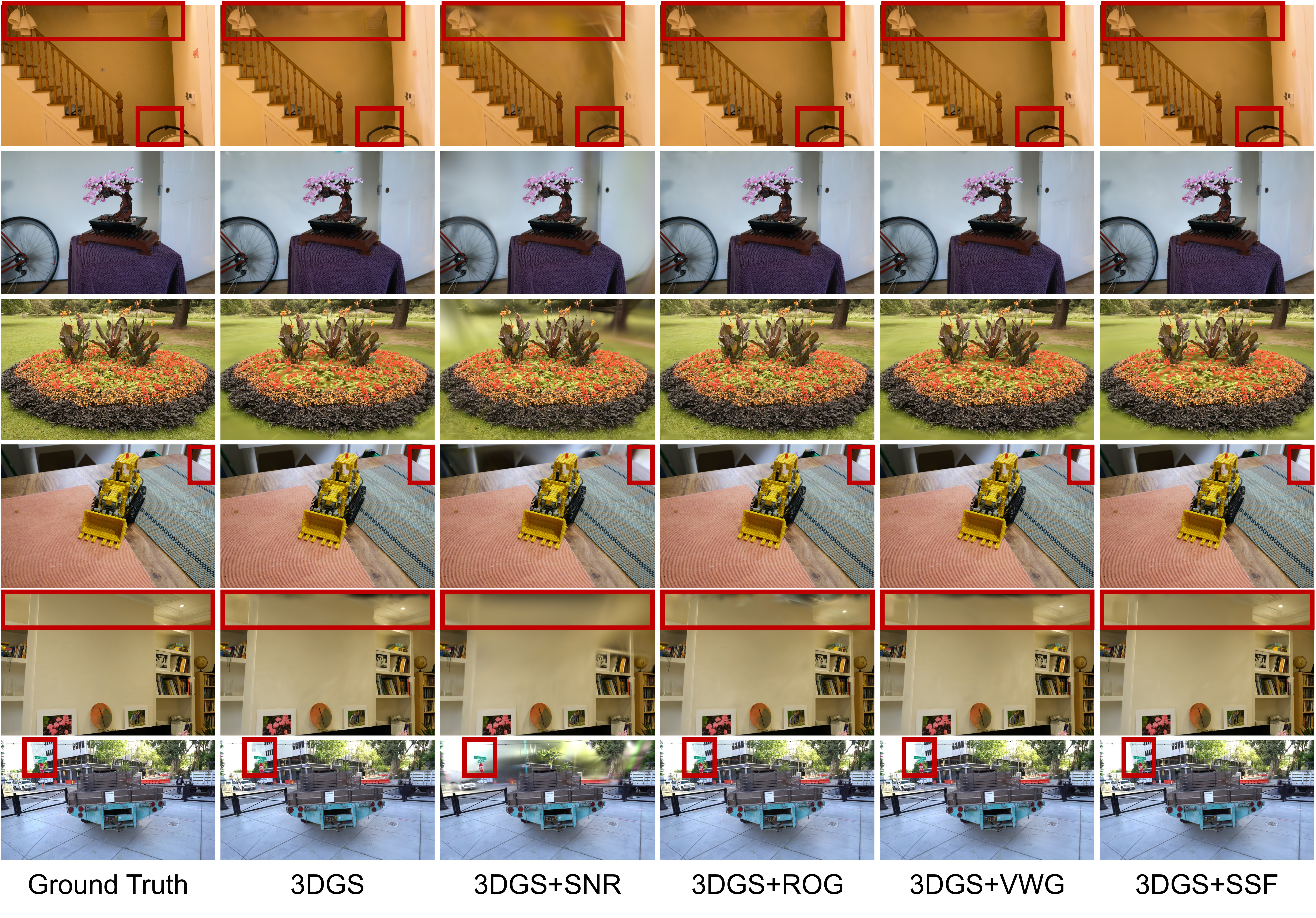}
    \caption{\textbf{Qualitative comparison.} SSF reduces Gaussian counts and
    accelerates training while maintaining rendering quality comparable to 3DGS
    and even reducing artifacts in regions such as the upper part of the
    \texttt{Room} scene, where unreliable densification signals lead to
    Gaussians being split or pruned before they adequately cover those regions.}
    \label{fig:fig5}
\end{figure*}

SNR substantially reduces quality across all three datasets. Its world-space
Adam-moment ratio measures directional consistency rather than image-space
capacity demand, a possible source of this loss. ROG yields the
largest densification and pruning counts and the slowest training, consistent
with the cost of maintaining a larger population and performing more topology
updates. VWG closely matches vanilla 3DGS because fixing temporal lifespan to one makes its selection rule equivalent to vanilla ADC, using the same mean gradient over visible observations and the same densification threshold.
\autoref{fig:fig5} compares matched held-out views and detail crops.

% On Tanks\&Temples, SSF reduces the Gaussian count by 18.2\%
% and training time by 6.2\%, with PSNR changing from
% 23.758 to 24.107\,dB.
% On Deep Blending, SSF reduces the Gaussian count by 19.0\%
% and training time by 14.3\%, with PSNR changing from
% 29.701 to 29.796\,dB.

\paragraph{SSF on Alternative Densification Signals.}
We evaluate SSF with AbsGS to test its applicability beyond vanilla signed
screen-space gradient accumulation. As shown in \autoref{tab:adc-variant},
SSF improves PSNR and SSIM across all three datasets, with PSNR increasing
from 27.001 to 27.196\,dB on MipNeRF360, from 23.678 to 23.906\,dB on
Tanks\&Temples, and from 29.634 to 29.756\,dB on Deep Blending.
Gaussian counts decrease by 33.8\%, 20.7\%, and 20.7\%, respectively,
while training time decreases by 5.9\%, 0.2\%, and 11.0\%.
Peak GPU memory also falls by 19.2\%--32.2\% across the datasets.
As with 3DGS+SSF, training-time savings are smaller than Gaussian-count
reductions. 
% LPIPS improves on Tanks\&Temples and Deep Blending but slightly
% worsens on MipNeRF360, despite its PSNR and SSIM gains.

% Source: 2. A100_absgs_rgdgs+ssf.xlsx; second sheet (dataset summaries).
\begin{table}[t]
    \centering
    \caption{Effect of SSF on AbsGS, averaged by dataset. Best values are bolded.}
    \label{tab:adc-variant}

    \setlength{\tabcolsep}{3pt}
    \resizebox{\linewidth}{!}{%
    \begin{tabular}{lrrrrrrr}
        \toprule
        Method & PSNR$\uparrow$ & SSIM$\uparrow$ & LPIPS$\downarrow$ &
        Time$\downarrow$ & FPS$\uparrow$ & PM.$\downarrow$ &
        \#G.$\downarrow$ \\
        \midrule
        \multicolumn{8}{l}{\textbf{\texttt{MipNeRF360}}~\cite{Barron2022Mipnerf360}} \\
        % Run: absgs_0
        AbsGS~\cite{Ye2024AbsGS} & 27.001 & 0.7962 & \textbf{0.2217} & 1506 & 137.0 & 6.00 &
        3.53 \\
        % Run: absgs_ncwv_0
        AbsGS + SSF & \textbf{27.196} & \textbf{0.8005} & 0.2248 & \textbf{1417} & \textbf{142.3} & \textbf{4.07} &
        \textbf{2.34} \\
        \midrule
        \multicolumn{8}{l}{\textbf{\texttt{Tanks\&Temples}}~\cite{Knapitsch2017Tanks}} \\
        % Run: absgs_0
        AbsGS~\cite{Ye2024AbsGS} & 23.678 & 0.8526 & 0.1623 & 586 & 223.4 & 2.18 &
        1.31 \\
        % Run: absgs_ncwv_0
        AbsGS + SSF & \textbf{23.906} & \textbf{0.8555} & \textbf{0.1617} & \textbf{585} & \textbf{231.6} & \textbf{1.76} &
        \textbf{1.04} \\
        \midrule
        \multicolumn{8}{l}{\textbf{\texttt{Deep Blending}}~\cite{Hedman2018Deep}} \\
        % Run: absgs_0
        AbsGS~\cite{Ye2024AbsGS} & 29.634 & 0.9010 & 0.2366 & 1009 & 204.3 & 3.75 &
        1.93 \\
        % Run: absgs_ncwv_0
        AbsGS + SSF & \textbf{29.756} & \textbf{0.9055} & \textbf{0.2344} & \textbf{898} & \textbf{215.6} & \textbf{2.66} &
        \textbf{1.53} \\
        \bottomrule
    \end{tabular}%
    }
\end{table}

\paragraph{Additional Training Acceleration.}
ImprovedGS, LeGS, FastGS, and AdpSplit accelerate training through different
mechanisms while retaining signal-based densification decisions. SSF
complements these mechanisms by deferring topology updates for Gaussians
whose signals have not settled, enabling
additional efficiency gains as shown in \autoref{tab:integration}.
With ImprovedGS, SSF improves all three rendering quality metrics on
MipNeRF360 and Tanks\&Temples, although training-time savings are modest.
These modest savings reflect ImprovedGS's Gaussian budgeting strategy,
which leaves little room for additional acceleration with SSF.
We therefore also evaluate its
unbudgeted variant, for which SSF reduces training time by 7.2--11.7\% and
Gaussian counts by 9.8--15.1\% across the three datasets, with PSNR increases
of 0.170--0.308\,dB.
A similar trend holds for LeGS and FastGS. On top of these baselines,
SSF reduces training time by 5.0--11.3\% for LeGS and by 5.5--11.2\% for FastGS. 
These gains accompany comparable or better PSNR, such as
increases from 28.280 to 28.318\,dB and from 27.913 to 27.976\,dB on
MipNeRF360 for LeGS+SSF and FastGS+SSF, respectively.
AdpSplit provides an integration with FastGS, and interestingly, SSF enables further
training-time reduction on top of it, reaching 250 seconds on MipNeRF360.
This corresponds to a $5.6\times$ speedup over vanilla 3DGS, with a negligible
PSNR degradation of $-0.025$\,dB relative to FastGS+AdpSplit.

% Source: 3. A100_legs_fastgs_adpsplit_improvedgs+ssf.xlsx; second sheet (dataset summaries).
\begin{table*}[t]
    \centering
    \caption{Additional training acceleration with SSF across efficient 3DGS backbones,
    averaged by dataset. UB denotes ImprovedGS without its Gaussian budget.
    Best values are bolded.}
    \label{tab:integration}
    \setlength{\tabcolsep}{3pt}
    \resizebox{\linewidth}{!}{%
    \begin{tabular}{l*{3}{rrrrrrr}}
        \toprule
        & \multicolumn{7}{c}{\textbf{\texttt{MipNeRF360}}~\cite{Barron2022Mipnerf360}}
        & \multicolumn{7}{c}{\textbf{\texttt{Tanks\&Temples}}~\cite{Knapitsch2017Tanks}}
        & \multicolumn{7}{c}{\textbf{\texttt{Deep Blending}}~\cite{Hedman2018Deep}} \\
        \cmidrule(lr){2-8}\cmidrule(lr){9-15}\cmidrule(lr){16-22}
        Method & PSNR$\uparrow$ & SSIM$\uparrow$ & LPIPS$\downarrow$ & Time$\downarrow$ & FPS$\uparrow$ & PM.$\downarrow$ & \#G.$\downarrow$ &
        PSNR$\uparrow$ & SSIM$\uparrow$ & LPIPS$\downarrow$ & Time$\downarrow$ & FPS$\uparrow$ & PM.$\downarrow$ & \#G.$\downarrow$ &
        PSNR$\uparrow$ & SSIM$\uparrow$ & LPIPS$\downarrow$ & Time$\downarrow$ & FPS$\uparrow$ & PM.$\downarrow$ & \#G.$\downarrow$ \\
        \midrule
        % Run: improvedgs_0
        ImprovedGS~\cite{Deng2026Improving} & 28.197 & 0.8350 & 0.1893 & 643 & \textbf{311.5} & \textbf{11.02} & 1.76 &
        24.331 & 0.8711 & 0.1461 & 408 & 376.1 & \textbf{4.90} & \textbf{1.25} &
        30.188 & 0.9141 & 0.2284 & \textbf{487} & \textbf{476.0} & \textbf{7.88} & \textbf{1.25} \\
        % Run: improvedgs_ncwv_0
        ImprovedGS + SSF & \textbf{28.263} & 0.8354 & 0.1883 & \textbf{641} & 303.7 & 11.06 & \textbf{1.74} &
        \textbf{24.665} & 0.8738 & 0.1433 & \textbf{384} & \textbf{394.8} & 4.92 & \textbf{1.25} &
        \textbf{30.197} & \textbf{0.9144} & 0.2290 & 491 & 469.5 & 7.92 & \textbf{1.25} \\
        % Run: improvedgs_nobdg_0
        ImprovedGS (UB) & 28.040 & 0.8355 & \textbf{0.1779} & 832 & 254.4 & 13.46 & 3.70 &
        24.330 & 0.8724 & 0.1404 & 461 & 328.3 & 5.60 & 1.73 &
        29.861 & 0.9095 & 0.2266 & 558 & 392.2 & 8.71 & 1.97 \\
        % Run: improvedgs_nobdg_ncwv_0
        ImprovedGS (UB) + SSF & 28.210 & \textbf{0.8367} & 0.1800 & 735 & 268.5 & 12.79 & 3.14 &
        24.567 & \textbf{0.8748} & \textbf{0.1392} & 410 & 351.8 & 5.38 & 1.56 &
        30.169 & 0.9141 & \textbf{0.2261} & 518 & 416.1 & 8.40 & 1.72 \\
        \midrule
        % Run: legs_0
        LeGS~\cite{Ning2026Beyond} & 28.280 & \textbf{0.8373} & \textbf{0.1831} & 518 & 327.1 & 11.29 & 2.16 &
        \textbf{24.809} & \textbf{0.8718} & \textbf{0.1429} & 460 & 298.7 & 9.49 & 1.67 &
        29.792 & 0.9063 & 0.2273 & 317 & 514.9 & 8.90 & 1.10 \\
        % Run: legs_fastgsncwv_0
        LeGS + SSF & \textbf{28.318} & \textbf{0.8373} & 0.1844 & \textbf{479} & \textbf{332.6} & \textbf{10.24} & \textbf{1.89} &
        24.734 & 0.8704 & 0.1479 & \textbf{408} & \textbf{308.6} & \textbf{8.49} & \textbf{1.42} &
        \textbf{30.091} & \textbf{0.9127} & \textbf{0.2258} & \textbf{301} & \textbf{515.2} & \textbf{8.05} & \textbf{1.08} \\
        \midrule
        % Run: fastgs_0
        FastGS~\cite{Ren2025FastGS} & 27.913 & \textbf{0.8190} & \textbf{0.2162} & 294 & 372.1 & 6.65 & 1.16 &
        \textbf{24.448} & \textbf{0.8549} & \textbf{0.1754} & 186 & 384.2 & 2.84 & 0.54 &
        30.170 & 0.9074 & \textbf{0.2431} & \textbf{172} & \textbf{486.8} & 4.65 & 0.65 \\
        % Run: fastgs_ncwv_0
        FastGS + SSF & \textbf{27.976} & 0.8177 & 0.2223 & \textbf{261} & \textbf{386.9} & \textbf{6.21} & \textbf{0.95} &
        24.254 & 0.8518 & 0.1797 & \textbf{176} & \textbf{421.2} & \textbf{2.78} & \textbf{0.49} &
        \textbf{30.182} & \textbf{0.9077} & 0.2441 & 173 & 480.8 & \textbf{4.54} & \textbf{0.58} \\
        \midrule
        % Run: adpsplit_fastgs_0
        FastGS + AdpSplit~\cite{Lee2026AdpSplit} & \textbf{27.964} & \textbf{0.8197} & \textbf{0.2166} & 271 & 505.3 & 7.56 & 1.23 &
        \textbf{24.388} & \textbf{0.8556} & \textbf{0.1747} & 170 & 598.9 & 3.19 & 0.55 &
        30.167 & 0.9099 & \textbf{0.2413} & 169 & 692.3 & 7.61 & 0.65 \\
        % Run: adpsplit_fastgs_ncwv_0
        FastGS + AdpSplit + SSF & 27.938 & 0.8168 & 0.2232 & \textbf{250} & \textbf{512.2} & \textbf{6.71} & \textbf{0.94} &
        24.298 & 0.8526 & 0.1794 & \textbf{159} & \textbf{635.4} & \textbf{2.95} & \textbf{0.48} &
        \textbf{30.253} & \textbf{0.9106} & 0.2432 & \textbf{148} & \textbf{775.0} & \textbf{5.37} & \textbf{0.53} \\
        \bottomrule
    \end{tabular}%
    }
\end{table*}

\begin{figure*}[t]
    \centering
    \includegraphics[width=.9\linewidth]{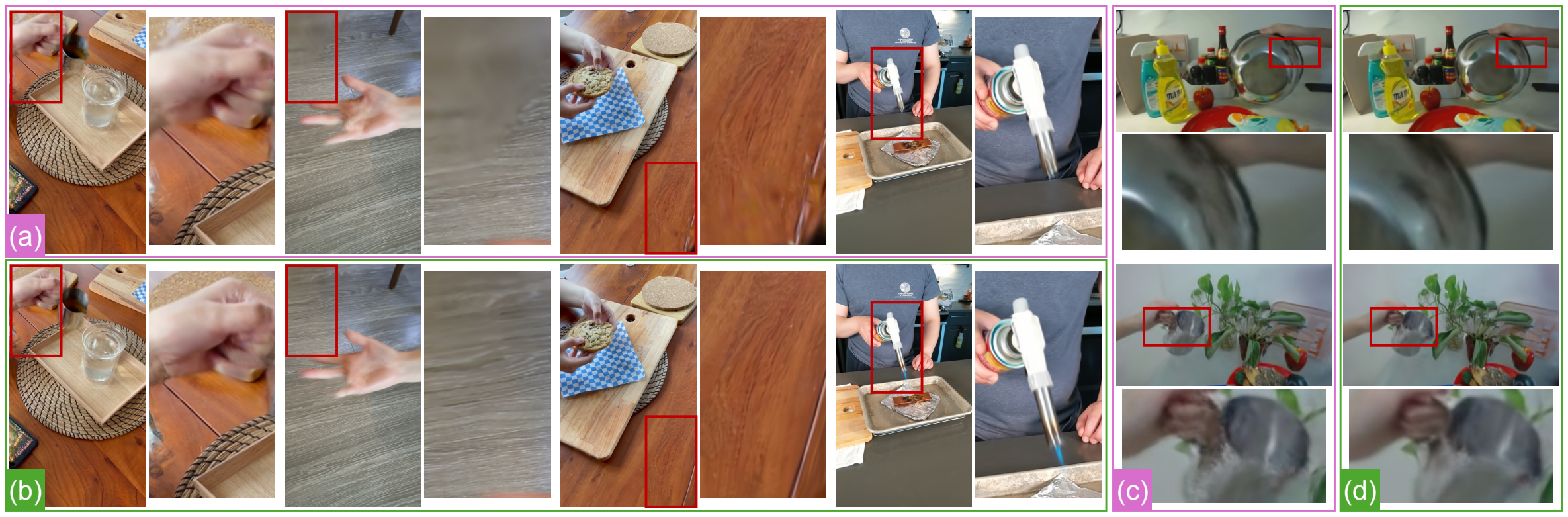}
    \caption{\textbf{Qualitative comparison on dynamic scenes.}
    On HyperNeRF, (a) Deformable-GS blurs moving objects and surrounding
    textures or misses details such as the torch flame, whereas
    (b) Deformable-GS+SSF preserves these details more faithfully.
    On NeRF-DS, (c) Deformable-GS exhibits distorted rigid-object contours,
    while (d) Deformable-GS+SSF reduces these artifacts.}
    \label{fig:fig6}
\end{figure*}

\begin{figure}[t]
    \centering
    \includegraphics[width=\linewidth]{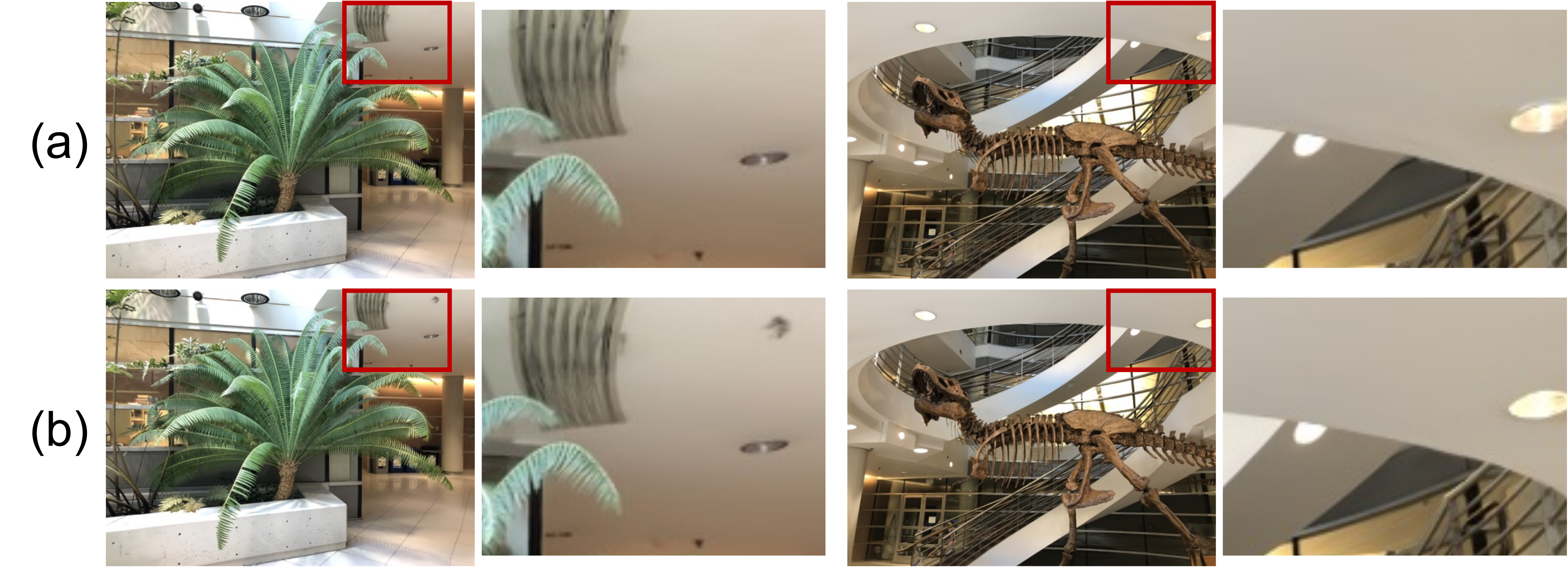}
    \caption{\textbf{Qualitative comparison on sparse-view scenes.}
    (a) DropGaussian misses small structures or renders them with artifacts,
    while (b) DropGaussian+SSF recovers the ceiling sprinkler and
    improves the appearance of the recessed light.}
    \label{fig:fig7}
\end{figure}

\paragraph{Generalization to Dynamic and Sparse-View Scenes.}
\label{sec:generalization-results}
Beyond densely observed static scenes, we evaluate SSF in more challenging
dynamic and sparse-view settings (\autoref{tab:generalization}). With
Deformable-GS, SSF improves PSNR, SSIM, and LPIPS on both HyperNeRF and
NeRF-DS, reducing training time by 31.4\% and 26.6\% and Gaussian counts
by 25.4\% and 23.1\%, respectively. Unlike the preceding comparisons,
training-time savings exceed the reduction in the final Gaussian count.
This is because Deformable-GS creates many redundant intermediate Gaussians
during early densification, as evidenced by its higher peak memory usage,
thereby increasing training time.
SSF also reduces blur and recovers details missing from the backbone's
renderings (\autoref{fig:fig6}). SSF allows Gaussians to learn deformation
more fully before densification and protects them from premature pruning
while their parameters are still being optimized, helping preserve geometry
and reduce artifacts in dynamic scenes.
For DropGaussian, SSF reduces training time by only two seconds (1.7\%)
but improves PSNR and SSIM, although LPIPS increases slightly.
The qualitative results show improved reconstruction of structures missed
by the backbone, including a ceiling sprinkler (\autoref{fig:fig7}).

% Source: 4. A100_dropgaussian_Deformable-GS+ssf.xlsx; second sheet (dataset summaries).
\begin{table}[t]
    \centering
    \caption{Generalization of SSF to dynamic reconstruction and sparse-view 
    training, averaged by dataset. Best values are bolded.}
    \label{tab:generalization}

    \setlength{\tabcolsep}{3pt}
    \resizebox{\linewidth}{!}{%
    \begin{tabular}{lrrrrrrr}
        \toprule
        Method & PSNR$\uparrow$ & SSIM$\uparrow$ & LPIPS$\downarrow$ &
        Time$\downarrow$ & FPS$\uparrow$ & PM.$\downarrow$ &
        \#G.$\downarrow$ \\
        \midrule
        \multicolumn{8}{l}{\textbf{\texttt{HyperNeRF}}~\cite{Park2021HyperNeRF}} \\
        % Run: Deformable-GS_0
        Deformable-GS~\cite{Yang2024Deformable} & 27.116 & 0.8387 & 0.1966 & 2305 & 20.2 & 12.84 &
        0.67 \\
        % Run: Deformable-GS_ncwv_0
        Deformable-GS + SSF & \textbf{27.249} & \textbf{0.8420} & \textbf{0.1882} & \textbf{1581} & \textbf{25.6} & \textbf{8.61} &
        \textbf{0.50} \\
        \midrule
        \multicolumn{8}{l}{\textbf{\texttt{NeRF-DS}}~\cite{Yan2023NeRFDS}} \\
        % Run: Deformable-GS_0
        Deformable-GS~\cite{Yang2024Deformable} & 23.804 & 0.8509 & 0.1802 & 662 & 77.0 & 4.08 &
        0.13 \\
        % Run: Deformable-GS_ncwv_0
        Deformable-GS + SSF & \textbf{23.825} & \textbf{0.8533} & \textbf{0.1758} & \textbf{486} & \textbf{96.2} & \textbf{3.33} &
        \textbf{0.10} \\
        \midrule
        \multicolumn{8}{l}{\textbf{\texttt{LLFF}}~\cite{Mildenhall2019Local} (9 views)} \\
        % Run: dropgaussian_v9_0
        DropGaussian~\cite{Park2025DropGaussian} & 26.078 & 0.8733 & \textbf{0.0890} & 119 & 341.5 & 0.82 &
        0.42 \\
        % Run: dropgaussian_v9_ncwv_0
        DropGaussian + SSF & \textbf{26.256} & \textbf{0.8739} & 0.0921 & \textbf{117} & \textbf{356.9} & \textbf{0.80} &
        \textbf{0.40} \\
        \bottomrule
    \end{tabular}%
    }
\end{table}

\paragraph{Ablation Study.}
\label{sec:ablation-results}
We ablate the components of SSF to assess their contributions relative to
both vanilla 3DGS and the full configuration. As shown in
\autoref{tab:ablation}, normalized signal change alone improves PSNR while
reducing the Gaussian count. Adding EMA smoothing further improves PSNR,
SSIM, and LPIPS, with a modest increase in Gaussian count. Combining
normalized signal change with consecutive-window validation reduces
cumulative densification and the final population further, but slightly
degrades reconstruction quality relative to normalized signal change alone.
The full configuration improves quality over this unsmoothed variant at a
small additional training cost, offering a balance between reconstruction
quality and training efficiency.

% ssf ablation
% 0. baseline: no use of --use_ncwv
% 1. low signal only: signal smoothing off, consecutive window validation off
% 2. smoothing low signal; while consecutive window validation off
% 3. use consecutive window validation; while using low signal
% 4. full component of ncwv

% | Case | Smoothing | Consecutive passes | Required change |
% |---|---|---:|---|
% | 0. Baseline | — | — | Omit --use_ncwv |
% | 1. Raw signal, no persistence | Off | 1 | Add a raw-signal smoother mode |
% | 2. Smoothed signal, no persistence | EMA | 1 | Set --NCWV_persistence_length 1 |
% | 3. Raw signal with persistence | Off | 3 | Add a raw-signal smoother mode |
% | 4. Full NCWV | EMA | 3 | Existing defaults |

% Source: 5. A100_ncwv_ablation.xlsx; baseline and full SSF from 1. A100_3dgs_cadam_rds_tad_vs_ssf.xlsx; second sheet (dataset summaries).
\begin{table}[t]
    \centering
    \caption{Component ablation on \texttt{MipNeRF360}. EMA denotes signal smoothing, NSC normalized signal change,
    and CWV consecutive-window validation. Best values are bolded, except for \#D and \#P.}
    \label{tab:ablation}

    \setlength{\tabcolsep}{3pt}
    \resizebox{\linewidth}{!}{%
    \begin{tabular}{lcccrrrrrrrrr}
        \toprule
        Method & EMA & NSC & CWV & PSNR$\uparrow$ & SSIM$\uparrow$ & LPIPS$\downarrow$ &
        Time$\downarrow$ & FPS$\uparrow$ & PM.$\downarrow$ &
        \#G.$\downarrow$ & \#D & \#P \\
        \midrule
        % Run: 3dgs_0
        3DGS~\cite{Kerbl20233DGaussian} &  &  &  & 27.526 & 0.8127 & \textbf{0.2208} & 1400 & 109.2 & 9.92 &
        2.64 & 3.13 & 0.61 \\
        \midrule
        % Run: ncwv_ablation-1.nsc
        NSC &  & $\checkmark$ &  & 27.590 & 0.8121 & 0.2254 & 1322 & 115.6 & 9.08 &
        2.05 & 2.30 & 0.37 \\
        % Run: ncwv_ablation-2.ema_nsc
        EMA + NSC & $\checkmark$ & $\checkmark$ &  & \textbf{27.631} & \textbf{0.8128} & 0.2243 & 1332 & 115.1 & 9.30 &
        2.19 & 2.51 & 0.43 \\
        % Run: ncwv_ablation-3.nsc_cwv
        NSC + CWV &  & $\checkmark$ & $\checkmark$ & 27.546 & 0.8100 & 0.2298 & \textbf{1286} & \textbf{119.9} & \textbf{8.61} &
        \textbf{1.73} & 1.90 & 0.28 \\
        \midrule
        % Run: 3dgs_ncwv_ema_tol0.05_w100_tout800_0
        3DGS + SSF & $\checkmark$ & $\checkmark$ & $\checkmark$ & 27.572 & 0.8108 & 0.2285 & 1292 & 118.8 & 8.75 &
        1.83 & 2.02 & 0.30 \\
        \bottomrule
    \end{tabular}%
    }
\end{table}

\section{Conclusion}
We presented SSF-GS, a steady-state filtering approach that addresses the
unreliability of transient densification signals in 3D Gaussian Splatting.
By combining signal smoothing, normalized signal change, and consecutive-window
validation, SSF allows Gaussian parameters to undergo further optimization
before the backbone determines whether to densify or prune them using its
existing selection criteria. Experiments demonstrate reduced Gaussian
populations and training costs, with reconstruction quality largely preserved
and improved in several settings. Integrations with alternative densification
signals, accelerated backbones, dynamic reconstruction, and sparse-view
training demonstrate its broad applicability. These findings support signal
stability as a practical consideration for Gaussian selection, complementary
to improvements in densification signals and operators.

\paragraph{Limitations.}
Signal stability does not ensure accurate geometry, so SSF offers only modest
quality gains on some backbones and datasets.
Furthermore, Gaussian-count reductions do not yield proportional training-time
savings, as continued optimization can enlarge projected footprints and
increase rendering cost per Gaussian. Finally, SSF requires signal-based
densification during iterative optimization, making it inapplicable to purely
feed-forward reconstruction unless followed by a refinement stage with such
density control.

\section*{Acknowledgments}
Supported by the Intelligence Advanced Research Projects Activity (IARPA) via Department of Interior/ Interior Business Center (DOI/IBC) contract number 140D0423C0076. The U.S. Government is authorized to reproduce and distribute reprints for Governmental purposes notwithstanding any copyright annotation thereon. Disclaimer: The views and conclusions contained herein are those of the authors and should not be interpreted as necessarily representing the official policies or endorsements, either expressed or implied, of IARPA, DOI/IBC, or the U.S. Government.

\section*{LLM Usage Statement}
We used OpenAI's Codex for coding assistance and language editing.
The authors reviewed the AI-assisted outputs and take responsibility for
the final manuscript and results.

{
    \small
    \bibliographystyle{ieeenat_fullname}
    \bibliography{main}
}

\clearpage
\setcounter{page}{1}
\maketitlesupplementary

\section{Experimental Setup for the Gradient Diagnosis}
\label{sec:diagnosis-setup}
We record the vanilla ADC mean screen-space position-gradient magnitude
$\bar g_i(t)$ every $T_d=100$ iterations and use a densification threshold
of $\tau_g=0.0002$, following the default settings of 3DGS.
Let $a_i(t)$ denote the number of iterations since Gaussian $i$ was created.
Every Gaussian is initially protected from cloning, splitting, and pruning
while $a_i(t)<1{,}000$.
Let $t_{i,\tau_g}$ denote the first training iteration at which Gaussian $i$
has a finite, observed window mean satisfying $\bar g_i(t)\geq\tau_g$.
At this iteration, we set its protection endpoint to age
$a_i(t_{i,\tau_g})+900$, so qualification at ages 100 and 900 gives
endpoints of 1,000 and 1,800, respectively.
Subsequent changes in threshold qualification do not extend protection.

For the reported \texttt{Bicycle} run, the cohort comprises newly created
Gaussians that first qualify at age 100 or later and complete their
900-iteration follow-up before densification ends at training iteration
15,000, satisfying $t_{i,\tau_g}+900<15{,}000$.
The cohort contains 790,221 Gaussians and excludes the 54,275 initial
Gaussians and 75,807 incomplete follow-ups at the densification end.
% Gaussians that complete follow-up remain in the statistics even if they
% are subsequently split or pruned.
We compute population statistics using this fixed cohort and align each
Gaussian's first threshold qualification to age 100, followed by nine
observations through aligned age 1,000.
% Missing or nonfinite observations retain the previous valid threshold
% decision when computing the population curves.

\section{Implementation of SSF and the Comparison Methods}
\label{sec:baseline-adaptations}

\subsection{3DGS+SSF Algorithm}
\label{sec:ncwv-algorithm}
\autoref{alg:ncwv} summarizes 3DGS training with SSF-gated density control.
Backbone 3DGS calls retain camera sampling, rendering, loss computation,
optimization, and opacity resets. Backbone ADC calls retain gradient
accumulation, candidate selection, topology operators, and accumulator resets.
SSF only gates densification and pruning candidates.
The normalized signal change (NSC) check period equals the densification interval $T_d=100$.
$t_{\mathrm{op\_reset}}=\{3k,6k,9k,12k\}$ denotes the set of opacity-reset iterations.

For each Gaussian, fresh SSF state consists of its birth iteration $b_i$,
zero observation counts $n_i,n_i^{\mathrm{prev}}$, EMA accumulator $m_{i,0}$,
peak $p_{i,0}$, and streak $s_i$, an unset reference
$\hat m_i^{\mathrm{prev}}=\bot$, and $\mathrm{settled}_i=\mathrm{false}$.
Here, $n_i^{\mathrm{prev}}$ records the observation count at the previous check,
and $a_i(t)=t-b_i$. The mask $e_i(t)$ is evaluated from the current state
whenever used. For brevity, numerical values are assumed finite and
$p_{i,n}>0$ when evaluating $d_i$.
Children receive fresh SSF state with $b_i=t$, clone parents retain theirs,
and removed Gaussians have their state discarded. 
% Backbone ADC resets and opacity resets
% do not reset SSF histories; opacity resets are not gated by SSF.

In the following four listings, an ADC round satisfies
$t_{\mathrm{densify\_from}}<t<t_{\mathrm{densify\_until}}$ and
$t\bmod T_d=0$. A backbone forward/backward call includes camera sampling,
rendering, loss computation, and backpropagation, with the usual learning-rate
and spherical-harmonic schedules. Backbone ADC calls include scale-based
cloning/splitting, pruning, Adam-state resizing, and accumulator resets.

\begin{algorithm}[t]
\small
\algrenewcommand{\algorithmicindent}{0.8em}
\caption{Pseudocode of 3DGS+SSF training}
\label{alg:ncwv}
\begin{algorithmic}[1]
\Require Training data, initial points, $T$, backbone settings
\Require $\beta,a_{\mathrm{warm}},a_{\mathrm{timeout}},\epsilon_{\mathrm{tol}},K$
\State Initialize backbone 3DGS and SSF ($b_i=0$)
\For{$t=1,\ldots,T$}
    \State Run backbone 3DGS forward/backward
    \If{$t<t_{\mathrm{densify\_until}}$}
        \State Update backbone ADC statistics; collect $g_i(t)$
        % \Comment{\autoref{eq:adc}}
        \For{each visible Gaussian $i$}
            \State $n_i\gets n_i+1$
            \State $m_{i,n_i}\gets\beta m_{i,n_i-1}+(1-\beta)g_i(t)$
            \Comment{\autoref{eq:ema}}
            \State $\hat m_{i,n_i}\gets m_{i,n_i}/(1-\beta^{n_i})$
            \Comment{\autoref{eq:ema}}
            \State $p_{i,n_i}\gets\max(p_{i,n_i-1},\hat m_{i,n_i})$
            \Comment{\autoref{eq:peak}}
        \EndFor
        \If{$t\bmod T_d=0$}
            \For{each $i$ with $e_i(t)=0$}
            \Comment{\autoref{eq:eligibility}}
                \If{$a_i(t)\geq a_{\mathrm{warm}}$ and $n_i>n_i^{\mathrm{prev}}$}
                    \If{$\hat m_i^{\mathrm{prev}}\neq\bot$}
                        \State $d_i\gets|\hat m_{i,n_i}-\hat m_i^{\mathrm{prev}}|/p_{i,n_i}$
                        \Comment{\autoref{eq:drift}}
                        \State $s_i(t)\gets(s_i(t)+1)\mathbb{1}[d_i\leq\epsilon_{\mathrm{tol}}]$
                    \Else
                        \State $s_i(t)\gets0$
                    \EndIf
                    \State $\hat m_i^{\mathrm{prev}}\gets\hat m_{i,n_i}$
                    \State $\mathrm{settled}_i\gets[s_i(t)\geq K]$
                \Else
                    \State $s_i(t)\gets0$; $\hat m_i^{\mathrm{prev}}\gets\bot$
                \EndIf
                \State $n_i^{\mathrm{prev}}\gets n_i$
            \EndFor
        \EndIf
        \If{$t$ is an ADC round}
            \State Get $\mathcal D(t)$ from backbone ADC
            % \Comment{\autoref{eq:vanilla-selection}}
            \State $\mathcal D'(t)\gets\{i\in\mathcal D(t):e_i(t)=1\}$
            \Comment{\autoref{eq:selection}}
            \State Backbone ADC: clone/split $\mathcal D'(t)$
            \State Initialize child SSF; discard split-parent SSF
            \State Get updated backbone ADC pruning set $\mathcal P(t)$
            % \Comment{\autoref{eq:vanilla-selection}}
            \State $\mathcal P'(t)\gets\{i\in\mathcal P(t):e_i(t)=1\}$
            \Comment{\autoref{eq:selection}}
            \State Backbone ADC: prune $\mathcal P'(t)$; discard SSF
            \State Reset backbone ADC statistics and $\mathcal W_i$
        \EndIf
        \If{$t\in t_{\mathrm{op\_reset}}$}
            \State Call backbone 3DGS opacity reset
        \EndIf
    \EndIf
    \If{$t<T$}
        \State Run backbone optimizer; clear gradients
    \EndIf
\EndFor
% \State \Return Trained Gaussians $\mathcal G$
\end{algorithmic}
\end{algorithm}

\subsection{3DGS+SNR Algorithm}
\autoref{alg:cadam} summarizes 3DGS training with SNR-based densification
and selective opacity reset.
Our CAdam adaptation~\cite{Chung2026CAdam} reads the first and second moments
$\mathbf{m}_i,\mathbf{v}_i$ stored by Adam for world-space positions.
It maintains a per-Gaussian count $u_i$ of completed visible position updates,
and uses the position optimizer's decay factors $\beta_1,\beta_2$ to compute
\begin{align}
    \hat{\mathbf{m}}_i &=
    \frac{\mathbf{m}_i}{\max(1-\beta_1^{u_i},b_{\min})},\\
    \hat{\mathbf{v}}_i &=
    \frac{\mathbf{v}_i}{\max(1-\beta_2^{u_i},b_{\min})},\\
    q_i &= \frac{\|\hat{\mathbf{m}}_i\|_2}
    {\sqrt{\|\hat{\mathbf{v}}_i\|_1}+\epsilon_{\mathrm{SNR}}}.
\end{align}
Here, $b_{\min}=10^{-3}$ is the bias-correction floor, and
$\epsilon_{\mathrm{SNR}}=10^{-9}$ is an additive numerical stabilizer.
At an ADC round, selection is $q_i>\tau_{\mathrm{SNR}}$ and replaces the ordinary
screen-gradient threshold rather than supplementing it. 
The reported MipNeRF360, Tanks\&Temples, and Deep-Blending runs 
use $0.50$, $0.60$, and $0.30$, respectively.

\begin{algorithm}[t]
\small
\caption{Pseudocode of 3DGS+SNR training}
\label{alg:cadam}
\begin{algorithmic}[1]
\Require Training data, initial points, $T$, backbone settings
\Require $\tau_{\mathrm{SNR}},\beta_1,\beta_2,b_{\min},\epsilon_{\mathrm{SNR}}$
\State Initialize backbone 3DGS; set $u_i\gets0$
\For{$t=1,\ldots,T$}
    \State Run backbone 3DGS forward/backward
    \If{$t<t_{\mathrm{densify\_until}}$}
        \State Update backbone ADC statistics
        \If{$t$ is an ADC round}
            \State Read Adam position moments $\mathbf m_i,\mathbf v_i$
            \State $\hat{\mathbf m}_i\gets\mathbf m_i/\max(1-\beta_1^{u_i},b_{\min})$
            \State $\hat{\mathbf v}_i\gets\mathbf v_i/\max(1-\beta_2^{u_i},b_{\min})$
            \State $q_i\gets\|\hat{\mathbf m}_i\|_2/(\sqrt{\|\hat{\mathbf v}_i\|_1}+\epsilon_{\mathrm{SNR}})$
            \State $\mathcal D\gets\{i:q_i>\tau_{\mathrm{SNR}}\}$
            \State Run backbone ADC with candidates $\mathcal D$
            \State Initialize child $u_i=0$; remove deleted Gaussians' $u_i$ entries
        \EndIf
        \If{$t\in t_{\mathrm{op\_reset}}$}
            \State Recompute $q_i$ from current position moments
            \State $\mathcal R\gets\{i:q_i<\tau_{\mathrm{SNR}},\ o_i>0.01\}$
            \State Set $o_i\gets0.01$ for $i\in\mathcal R$
            \State Zero opacity moments and gradients on $\mathcal R$
        \EndIf
    \EndIf
    \If{$t<T$}
        \State Run backbone 3DGS optimizer
        \If{$t<t_{\mathrm{densify\_until}}$ and position gradients exist}
            \State Increment $u_i$ for visible Gaussians
        \EndIf
        \State Clear gradients
    \EndIf
\EndFor
\end{algorithmic}
\end{algorithm}

\subsection{3DGS+ROG Algorithm}
\autoref{alg:rds} summarizes 3DGS training with observation-weighted
gradient statistics for densification.
We adapt the observation-weighting method of RDS-GS~\cite{Pang2026Improving}
to the backbone's single-view sampling and optimization, rather than the
original method's joint optimization over two views per iteration.
For the sampled view at iteration $t$, $m_i(t)$ counts the pixels to which
Gaussian $i$ contributes during alpha compositing.
We use $\omega_i(t)=\sqrt{m_i(t)}$ as a single-view adaptation.
Here, $m_i(t)$ is distinct from the SSF EMA accumulator $m_{i,n}$.
We accumulate the weighted screen-space responses across iterations within
each densification interval:
\begin{equation}
    \bar g_i^{\mathrm{RDS}}=
    \frac{\sum_t \omega_i(t)g_i(t)}{\sum_t \omega_i(t)+\epsilon_{\mathrm{ROG}}}.
\end{equation}
Here, $\epsilon_{\mathrm{ROG}}=10^{-6}$ is an additive numerical stabilizer.
Selection uses the strict comparison $\bar g_i^{\mathrm{RDS}}>\tau_g$ ($\tau_g=0.0002$ in default).

\begin{algorithm}[t]
\small
\caption{Pseudocode of 3DGS+ROG training}
\label{alg:rds}
\begin{algorithmic}[1]
\Require Training data, initial points, $T$, backbone settings
\Require Densification threshold $\tau_g$
\State Initialize backbone 3DGS; set $A_i,B_i\gets0$
\For{$t=1,\ldots,T$}
    \State Run backbone 3DGS forward/backward
    \If{$t<t_{\mathrm{densify\_until}}$}
        \State Collect gradients $g_i(t)$ and pixel counts $m_i(t)$
        \For{each Gaussian $i$}
            \State $\omega_i(t)\gets\sqrt{m_i(t)}$
            \State $A_i\gets A_i+\omega_i(t)g_i(t)$; $B_i\gets B_i+\omega_i(t)$
        \EndFor
        \State Update backbone ADC radius statistics
        \If{$t$ is an ADC round}
            \State $\bar g_i^{\mathrm{RDS}}\gets A_i/(B_i+\epsilon_{\mathrm{ROG}})$
            \State $\mathcal D\gets\{i:\bar g_i^{\mathrm{RDS}}>\tau_g\}$
            \State Run backbone ADC with candidates $\mathcal D$
            \State Reset $A_i,B_i$ for all live Gaussians
        \EndIf
        \If{$t\in t_{\mathrm{op\_reset}}$}
            \State Call backbone 3DGS opacity reset
        \EndIf
    \EndIf
    \If{$t<T$}
        \State Run backbone optimizer; clear gradients
    \EndIf
\EndFor
\end{algorithmic}
\end{algorithm}

\subsection{3DGS+VWG Algorithm}
\autoref{alg:tad} summarizes 3DGS training with the fixed-lifespan
adaptation of temporally aware densification~\cite{Sandu2026Temporally}.
Its temporal weight $\sigma_i(u)$ denotes the Gaussian's opacity at frame $u$.
Our static adaptation uses unit weights $\sigma_i(u)=1$ for observations
$u\in\mathcal W_i$ and fixes $\psi_i=\bar\psi_i=1$.
The weighted mean therefore reduces to the visible-observation mean, and
the adjusted threshold reduces to $\tau_g$
with $\alpha_{\mathrm{TAT}}=1$ and $\beta_{\mathrm{TAT}}=0.3$.

\begin{algorithm}[t]
\small
\caption{Pseudocode of 3DGS+VWG training}
\label{alg:tad}
\begin{algorithmic}[1]
\Require Training data, initial points, $T$, backbone settings
\Require $\tau_g$, fixed $\psi_i=1$, $\alpha_{\mathrm{TAT}}=1$, $\beta_{\mathrm{TAT}}=0.3$
\State Initialize backbone 3DGS
\For{$t=1,\ldots,T$}
    \State Run backbone 3DGS forward/backward
    \If{$t<t_{\mathrm{densify\_until}}$}
        \State Update backbone ADC gradient/radius statistics
        \If{$t$ is an ADC round}
            \State $\sigma_i(u)\gets1$ for $u\in\mathcal W_i$
            \State $\bar g_i(t)\gets
            \frac{\sum_{u\in\mathcal W_i}\sigma_i(u)g_i(u)}
            {\sum_{u\in\mathcal W_i}\sigma_i(u)}$
            \State Use $\bar g_i(t)=0$ if $\mathcal W_i=\varnothing$
            \State $\bar\psi_i\gets1$
            \State $\tau_{g,i}\gets\tau_g[1+\beta_{\mathrm{TAT}}(1-\bar\psi_i)]^{-\alpha_{\mathrm{TAT}}}$
            \State $\mathcal D\gets\{i:\bar g_i(t)\geq\tau_{g,i}\}$
            \State Run backbone ADC with candidates $\mathcal D$
        \EndIf
        \If{$t\in t_{\mathrm{op\_reset}}$}
            \State Call backbone 3DGS opacity reset
        \EndIf
    \EndIf
    \If{$t<T$}
        \State Run backbone optimizer; clear gradients
    \EndIf
\EndFor
\end{algorithmic}
\end{algorithm}

\section{Hyperparameter Sweep}
\label{sec:hyperparameter-sensitivity}
\autoref{tab:hyperparameter-sensitivity} reports sensitivity to the EMA decay
$\beta$, timeout age $a_{\mathrm{timeout}}$, settling tolerance
$\epsilon_{\mathrm{tol}}$, persistence length $K$, and warmup age
$a_{\mathrm{warm}}$ on MipNeRF360~\cite{Barron2022Mipnerf360}.
We vary one parameter at a time around the default configuration
$\beta=0.999$, $a_{\mathrm{timeout}}=800$, $\epsilon_{\mathrm{tol}}=0.05$,
$K=3$, and $a_{\mathrm{warm}}=100$.
Each result is averaged over all nine scenes.
We use NVIDIA A100 GPU for these experiments.

Changing $\beta$ from 0.9 to 0.9999 yields PSNR values between 27.526 and
27.595 dB, with Gaussian counts between 1.74M and 1.83M.
Timeout age $a_{\mathrm{timeout}}$ more directly controls 
the tradeoff between representation size
and rendering quality. Increasing it from 400 to 1,200 iterations reduces
Gaussian count from 2.15M to 1.70M and training time from 1,372 to 1,252 seconds.
PSNR changes from 27.627 to 27.549 dB, while SSIM decreases from 0.8124
to 0.8097 and LPIPS increases from 0.2248 to 0.2307.
Allowing a longer wait suppresses more Gaussian growth but does not
uniformly preserve all quality metrics.

Settling tolerance $\epsilon_{\mathrm{tol}}$ and persistence length $K=3$
exhibit complementary trends.
Relaxing $\epsilon_{\mathrm{tol}}$ from 0.01 to 0.10 increases the population
from 1.71M to 1.95M, while improving PSNR from 27.512 to 27.583 dB.
Increasing $K$ from 1 to 5 instead reduces the population from 2.20M to
1.73M, with PSNR changing from 27.593 to 27.528 dB.

Warmup ages $a_{\mathrm{warm}}$ of 100--300 iterations have 
an even smaller effect, with PSNR
between 27.567 and 27.579 dB and counts between 1.79M and 1.83M.
These results indicate limited sensitivity to these parameters within the
evaluated ranges.

The default configuration provides an intermediate quality--efficiency
tradeoff rather than maximizing every metric.

\begin{table}[t]
\centering
\caption{Hyperparameter sensitivity of 3DGS+SSF, averaged over the nine
\texttt{MipNeRF360} scenes.
$\dagger$ marks the default configuration.
Time is measured in seconds, PM. denotes peak GPU memory in GB, and
\#G. denotes the final Gaussian count in millions. Ages are measured in
iterations. The best displayed value within each sweep is bolded.}
\label{tab:hyperparameter-sensitivity}
\resizebox{\linewidth}{!}{%
\begin{tabular}{llrrrrrrr}
\toprule
Parameter & Value & PSNR $\uparrow$ & SSIM $\uparrow$ & LPIPS $\downarrow$
& Time $\downarrow$ & FPS $\uparrow$ & PM. $\downarrow$ & \#G. $\downarrow$ \\
\midrule
$\beta$ & 0.9 & 27.528 & 0.8101 & 0.2300 & \textbf{1284} & 118.5 & \textbf{8.61} & \textbf{1.74} \\
 & 0.99 & 27.526 & 0.8107 & 0.2288 & 1291 & \textbf{119.1} & 8.72 & 1.81 \\
 & 0.999$^{\dagger}$ & 27.572 & \textbf{0.8108} & 0.2285 & 1292 & 118.8 & 8.75 & 1.83 \\
 & 0.9999 & \textbf{27.595} & \textbf{0.8108} & \textbf{0.2283} & 1320 & 116.9 & 8.75 & 1.83 \\
\midrule
$a_{\mathrm{timeout}}$ & 400 & \textbf{27.627} & \textbf{0.8124} & \textbf{0.2248} & 1372 & 112.3 & 9.24 & 2.15 \\
 & 500 & 27.570 & 0.8119 & 0.2258 & 1319 & 116.2 & 9.08 & 2.04 \\
 & 600 & 27.575 & 0.8116 & 0.2267 & 1327 & 115.5 & 8.95 & 1.96 \\
 & 700 & 27.523 & 0.8112 & 0.2274 & 1296 & 118.6 & 8.86 & 1.90 \\
 & 800$^{\dagger}$ & 27.572 & 0.8108 & 0.2285 & 1292 & 118.8 & 8.75 & 1.83 \\
 & 900 & 27.552 & 0.8106 & 0.2287 & 1288 & 119.1 & 8.69 & 1.79 \\
 & 1000 & 27.556 & 0.8104 & 0.2295 & 1280 & 119.8 & 8.63 & 1.75 \\
 & 1100 & 27.548 & 0.8100 & 0.2301 & 1273 & 120.2 & 8.59 & 1.72 \\
 & 1200 & 27.549 & 0.8097 & 0.2307 & \textbf{1252} & \textbf{121.9} & \textbf{8.55} & \textbf{1.70} \\
\midrule
$\epsilon_{\mathrm{tol}}$ & 0.01 & 27.512 & 0.8100 & 0.2301 & \textbf{1266} & \textbf{121.7} & \textbf{8.58} & \textbf{1.71} \\
 & 0.03 & 27.547 & 0.8107 & 0.2290 & 1284 & 119.4 & 8.66 & 1.77 \\
 & 0.05$^{\dagger}$ & 27.572 & 0.8108 & 0.2285 & 1292 & 118.8 & 8.75 & 1.83 \\
 & 0.07 & 27.577 & 0.8111 & 0.2275 & 1287 & 117.9 & 8.85 & 1.90 \\
 & 0.10 & \textbf{27.583} & \textbf{0.8118} & \textbf{0.2268} & 1309 & 116.3 & 8.94 & 1.95 \\
\midrule
$K$ & 1 & \textbf{27.593} & \textbf{0.8125} & \textbf{0.2242} & 1340 & 113.1 & 9.30 & 2.20 \\
 & 2 & \textbf{27.593} & 0.8116 & 0.2267 & 1297 & 117.4 & 8.93 & 1.95 \\
 & 3$^{\dagger}$ & 27.572 & 0.8108 & 0.2285 & 1292 & 118.8 & 8.75 & 1.83 \\
 & 4 & 27.536 & 0.8105 & 0.2290 & \textbf{1266} & 119.5 & 8.66 & 1.77 \\
 & 5 & 27.528 & 0.8103 & 0.2297 & 1271 & \textbf{121.1} & \textbf{8.60} & \textbf{1.73} \\
\midrule
$a_{\mathrm{warm}}$ & 100$^{\dagger}$ & 27.572 & 0.8108 & 0.2285 & 1292 & 118.8 & 8.75 & 1.83 \\
 & 200 & \textbf{27.579} & \textbf{0.8109} & \textbf{0.2283} & \textbf{1279} & 118.9 & 8.73 & 1.82 \\
 & 300 & 27.567 & 0.8106 & 0.2289 & 1280 & \textbf{120.0} & \textbf{8.69} & \textbf{1.79} \\
\bottomrule
\end{tabular}%
}
\end{table}

\end{document}